\pdfoutput=1

\documentclass[11pt]{article}

\usepackage[preprint]{acl}

\usepackage{times}
\usepackage{latexsym}
\usepackage{xcolor}
\usepackage[normalem]{ulem}  
\usepackage{soul}            

\usepackage[T1]{fontenc}

\usepackage[utf8]{inputenc}

\usepackage{microtype}

\usepackage{inconsolata}

\usepackage{graphicx}
\usepackage{enumitem}
\usepackage{multirow}
\usepackage{booktabs}
\usepackage{csquotes}
\usepackage{array}
\usepackage{amsmath}
\usepackage{siunitx} 
\usepackage{caption} 
\usepackage{todonotes}
\usepackage{listings}
\usepackage{xspace}
\usepackage{rotating}
\usepackage{colortbl}
\usepackage{float}
\usepackage{cleveref}
\usepackage{chngcntr}
\usepackage{multicol} 
\usepackage{caption}   
\usepackage{tikz} 
\usepackage{cuted}

\definecolor{mylightyellow}{rgb}{1.0, 0.98, 0.8}

\definecolor{lightgray}{rgb}{0.95,0.95,0.95} 
\definecolor{mylightgreenRGB}{RGB}{144, 238, 144}

\title{
\raisebox{-1.1ex}{\protect\includegraphics[trim={280 380 300 150}, clip, height=2.5\fontcharht\font`\B]{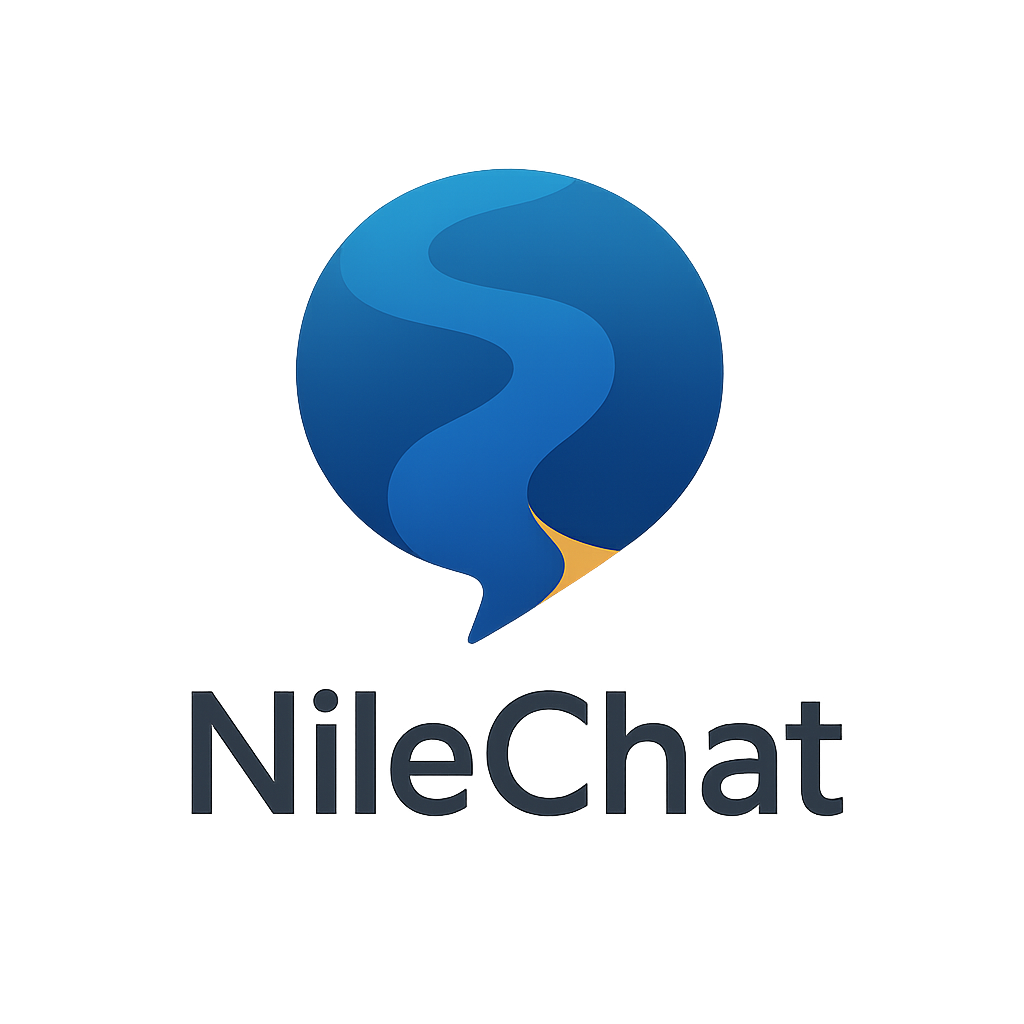}}
NileChat: Towards Linguistically Diverse and Culturally Aware LLMs for Local Communities}

\author{
    \normalsize \bfseries Abdellah {El Mekki}$^{\lambda}$\thanks{Work initiated during a stay at MBZUAI.} ~~~~~~~~
    \bfseries Houdaifa Atou$^{\delta}$ ~~~~~~~~
    \bfseries Omer Nacar$^{\psi}$ \\ 
    \normalsize \bfseries Shady Shehata$^{\gamma}$ ~~~~~~~~ 
    \bfseries Muhammad Abdul-Mageed$^{\lambda,\gamma}$ \\[1ex] 
    $^{\lambda}$The University of British Columbia ~~~~~~~~~
    $^{\delta}$Mohammed VI Polytechnic University ~~~~~~~~~ \\[1ex]
    $^{\psi}$Tuwaiq Academy  ~~~~~~~~~
    $^{\gamma}$Invertible AI \\[1ex]
    \texttt{ \{abdellah.elmekki,muhammad.mageed\}@ubc.ca}
}

\begin{document}
\maketitle

\begin{strip}
  \centering
  \includegraphics[width=\textwidth]{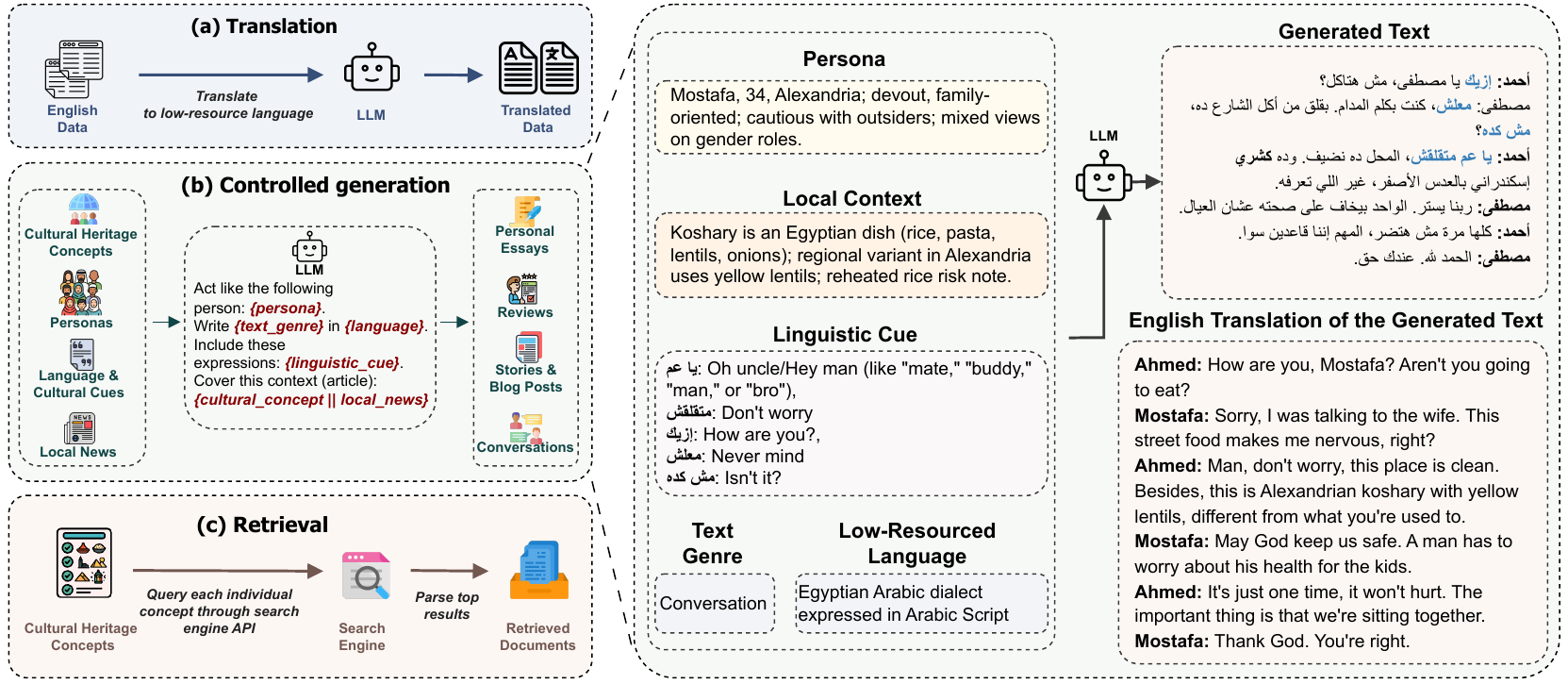}
  \captionof{figure}{Our proposed framework enhances text data augmentation for low-resource local communities through a multi-stage pipeline. First, it \textbf{(a)} generates educational data using \textbf{machine translation}. Next, it \textbf{(b)} creates diverse, culturally-aware texts, such as stories and conversations, by simulating scenarios with local personas through \textbf{controlled synthetic data generation}. Finally, it \textbf{(c)} enriches the model with local knowledge by \textbf{retrieving and parsing culturally specific web content}. This entire process enables controlled text generation and retrieval-augmented pre-training, ensuring the cultural and value alignment of large language models for Arabic dialects.}
  \label{fig:synthetic_pre_training}
\end{strip}

\begin{abstract}
Enhancing the linguistic capabilities of Large Language Models (LLMs) to include low-resource languages is a critical research area. Current research directions predominantly rely on synthetic data generated by translating English corpora, which, while demonstrating promising linguistic understanding and translation abilities, often results in models aligned with source language culture. These models frequently fail to represent the cultural heritage and values of local communities. This work proposes a methodology to create both synthetic and retrieval-based pre-training data tailored to a specific community, considering its \textit{(i) language}, \textit{(ii) cultural heritage}, and \textit{(iii) cultural values}. We demonstrate our methodology using Egyptian and Moroccan dialects as testbeds, chosen for their linguistic and cultural richness and current underrepresentation in LLMs. As a proof-of-concept, we develop \textit{NileChat}, a 3B parameter Egyptian and Moroccan Arabic LLM adapted for Egyptian and Moroccan communities, incorporating their language, cultural heritage, and values. Our results on various understanding, translation, and cultural and values alignment benchmarks show that \textit{NileChat} outperforms existing Arabic-aware LLMs of similar size and performs on par with larger models. 
This work addresses Arabic dialect in LLMs with a focus on cultural and values alignment via controlled synthetic data generation and retrieval-augmented pre-training for Moroccan Darija and Egyptian Arabic, including Arabizi variants, advancing Arabic NLP for low-resource communities.
We share our methods, data, and models with the community to promote the inclusion and coverage of more diverse communities in cultural LLM development.\footnote{\href{https://github.com/UBC-NLP/nilechat}{https://github.com/UBC-NLP/nilechat}.}

\end{abstract}

\section{Introduction}

Large Language Models (LLMs) have  advanced rapidly, enabling remarkable proficiency across many tasks. Yet, this success is unevenly distributed across languages, with substantial performance disparities observed for non-English languages, particularly low-resource languages and dialectal variants~\cite{10.1145/3597307}. A primary factor underlying this discrepancy is the limited representation of diverse multilingual data within the foundational pre-training corpora of these models which favors high-resource languages spoken in regions with high economic influence \cite{Bender_2011,joshi-etal-2020-state}.

Beyond linguistic limitations, a more profound challenge is the inherent risk of \textit{cultural encapsulation} \cite{10.1037/13132-000} in LLMs. Cultural encapsulation refers to an unconscious tendency to operate within one's own cultural lens, leading to misunderstanding or avoidance of differing perspectives and values. As LLMs are optimized to replicate patterns in their training data—predominantly sourced from specific cultural contexts (e.g., Western, English-speaking)—they risk internalizing and propagating these dominant perspectives as the norm \cite{dwivedi-etal-2023-eticor,10.1093/pnasnexus/pgae346, wang-etal-2024-countries, naous-etal-2024-beer}. The significance of cultural context cannot be overstated. As Edward Sapir noted:

\begin{displayquote}
\textit{"No two languages are ever sufficiently similar to be considered as representing the same social reality. The worlds in which different societies live are distinct worlds, not merely the same world with different labels attached." - \citet{Sapir1929}}
\end{displayquote}

This cultural bias is compounded by a fundamental mismatch: LLMs typically process data through a language-centric lens, whereas human communities are structured around shared social ties, perspectives, and values \cite{MacQueen2001}. Current LLMs adaptation techniques for new languages or communities \cite{gurgurov-etal-2024-adapting,joshi-etal-2025-adapting} often fall short in bridging this cultural divide, especially for low-resource communities \cite{naous-etal-2024-beer}. For instance, machine translation, while useful for generating synthetic data to boost linguistic coverage \cite{joshi-etal-2025-adapting, shang-etal-2025-atlas,wang-etal-2025-language}, primarily addresses the linguistic deficit. The translated content often retains the source language's cultural perspective, failing to incorporate authentic local nuances crucial for genuine interaction. Supervised fine-tuning (SFT) on target language data \cite{gala2024airavataintroducing, shang-etal-2025-atlas} can align models to specific tasks, but small datasets may not reshape deep-seated cultural biases from pre-training \cite{rystrøm2025multilingualmulticulturalevaluating} and can encourage hallucination with new factual data \cite{gekhman-etal-2024-fine}. While continued pre-training with culturally rich data could mitigate these issues, it faces a critical bottleneck for low-resource contexts: the scarcity of such high-quality digital texts.

This paper addresses the critical need to adapt multilingual LLMs to low-resource language communities by jointly considering their linguistic characteristics and cultural heritage \& values. We propose a novel pipeline (illustrated in Figure~\ref{fig:synthetic_pre_training}) focused on data augmentation for continued pre-training. Our approach combines \textit{controlled synthetic data generation} (Section \ref{sec:values_aware_pt_data}) with \textit{retrieval} (Section \ref{sec:retrieval_pt_data}) methods. To address linguistic adaptation, we translate English pre-training data into the target local language focusing only on high-quality data from the educational domain (Section \ref{sec:mt_pt_data}). Crucially, to imbue cultural relevance, we generate diverse texts reflecting specific cultural heritage concepts (e.g., food, celebrations, proverbs) using local persona descriptions (Section \ref{sec:values_aware_pt_data}) reflecting the local cultural values. We demonstrate our method on the \textit{Moroccan} and \textit{Egyptian} Arabic dialects as low-resource testbeds. We further pre-train a multilingual LLM on a curated mix of real and synthetic data, evaluating its performance on tasks involving language understanding, translation, and alignment with cultural knowledge and values. Our findings show that the adapted model significantly outperforms baseline and existing models that are even bigger in size on most evaluation tasks.

The main contributions of this work are: \textbf{(i)} A novel framework for augmenting pre-training corpora tailored to local communities. This framework considers their unique linguistic features, cultural heritage, and values by leveraging a teacher LLM. \textbf{(ii)} The public release of new datasets, representing the largest publicly available corpora for Egyptian and Moroccan Arabic dialects. These resources are intended to foster further research in these under-resourced languages. \textbf{(iii)} The development and public release of \textit{NileChat}, a robust 3-billion parameter LLM. This model demonstrates proficiency in both Egyptian and Moroccan dialectal Arabic (using Arabic script and Arabizi) while maintaining strong performance in Modern Standard Arabic, French, and English.



\section{Related Work}








\paragraph{Adaptation of LLMs.}
LLMs, despite general strengths, often require adaptation for specific languages, domains, or cultures \cite{bang-etal-2023-multitask, alkhamissi-etal-2024-investigating, naous-etal-2024-beer, song2025injectingdomainspecificknowledgelarge}. Adaptation techniques include prompt engineering \cite{shen-etal-2024-understanding}, SFT on culturally specific datasets \cite{huang-etal-2024-acegpt}, and continued pre-training on target-specific data \cite{fujii2024continual, huang-etal-2024-acegpt}. A key challenge, especially for SFT-based cultural adaptation, is the scarcity of comprehensive cultural datasets, hindering alignment with under-represented communities \cite{ahmad-etal-2024-generative, shen-etal-2024-understanding}.

\paragraph{Synthetic Data Augmentation for LLMs.}
To address data limitations, synthetic data augmentation has shown promise in improving LLM performance \cite{ge2024scaling, li2024culturellm, joshi-etal-2025-adapting}. Machine-translated data, for instance, can enhance capabilities in new languages \cite{joshi-etal-2025-adapting, shang-etal-2025-atlas}, and persona-driven synthetic data generation has also yielded performance gains \cite{ge2024scaling} and aided in tasks like assessing LLM political alignment \cite{bernardelle2024mapping}. However, synthetic data can sometimes degrade performance \cite{seddik2024how}, necessitating best practices for its use \cite{liu2024best}.

\paragraph{Arabic LLMs.}
In Arabic LLM development, models are either trained from scratch \cite{billah-nagoudi-etal-2023-jasmine, sengupta2023jais} or adapted from existing ones \cite{huang-etal-2024-acegpt, bari2025allam, fanarteam2025fanararabiccentricmultimodalgenerative}. A common method involves translating English data to Arabic, which, however, can introduce cultural biases from the source language \cite{sengupta2023jais, naous-etal-2024-beer}. Recent work on dialectal Arabic, such as translating instructions into Moroccan dialect for SFT, has improved generation tasks \cite{shang-etal-2025-atlas}. Yet, enhanced performance on standard tasks does not guarantee cultural awareness. While models like AceGPT \cite{huang-etal-2024-acegpt} and Fanar \cite{fanarteam2025fanararabiccentricmultimodalgenerative} aim for cultural cognizance, our work uniquely focuses on adapting existing LLMs to a local community by deeply integrating its specific linguistic features, cultural heritage, and values, building upon these prior advancements.

\section{Methodology}
In this work, we investigate the potential of pre-training data to imbue LLMs with the specific local characteristics of under-represented communities. We conceptualize these characteristics along three primary dimensions \cite{geertz1977interpretation, 1123_culture, bourdieu1991language, higgins2020communities, stanlaw2025language}: \textit{ (i) Language:} Encompassing dialectal nuances, idiomatic expressions, and linguistic structures unique to the community. \textit{(ii) Cultural Heritage:} Reflecting the customs, traditions, social norms, historical context, and common knowledge prevalent within the community. \textit{(iii) Cultural Values:} Capturing the ethical standpoints, belief systems, and societal priorities that define the community. We refer to these three dimensions as Language-Heritage-Values dimensions, \textit{LHV} for short. While we do not posit these as exhaustive of the attributes of a given community, we employ them as a vehicle to approximate the LLM communication and information needs at local levels. To ground our investigation, we focus on two low-resource varieties of Arabic—The Egyptian Arabic (EGY) and Moroccan Arabic (MOR)—. These dialects serve as our primary case studies for evaluating the methods proposed herein.

\subsection{Data Augmentation} \label{sec:data_aug}
The construction of linguistically-rich and culturally-rich LLMs that can serve a specific population fundamentally depends on the availability of representative data. Recognizing the acute scarcity of publicly available pre-training corpora for many low-resource languages, including EGY and MOR, we propose a novel data production method encapsulating the \textit{LHV} dimensions of a given country-level population. As depicted in Figure \ref{fig:synthetic_pre_training}, our approach leverages three complementary strategies intended to collectively capture the \textit{LHV} dimensions: \textit{(a) machine translation (MT)}, \textit{(b) controlled synthetic data generation} and \textit{(c) retrieval}. We explain these next.

\subsubsection{MT for Knowledge and Fluency}\label{sec:mt_pt_data}
To ensure linguistic fluency and coherence, we translate structured educational content from English into the target low-resource language using a specialized teacher model. Our pipeline preserves original formatting and includes filtering to remove unreliable translations identified by repetitive n-grams. We use educational materials for their topical breadth (including subjects such as education, history, health, medicine, and biology).


\subsubsection{Controlled Synthetic Data Generation for Cultural Heritage and Cultural Values} \label{sec:values_aware_pt_data} 

Linguistic fluency, while a foundational capability for LLMs, does not inherently guarantee their awareness of, or alignment with, the culture and values of a specific target community \cite{naous-etal-2024-beer}. To bridge this gap, we employ \textit{controlled synthetic data generation}. For controlled generation, we use the teacher LLM to generate diverse texts in the target language. These texts are specifically designed to discuss local topics, which are identified from articles sourced from local news websites or the target country's Wikipedia portal. Furthermore, the generated content is crafted to reflect distinct personas, each defined by a profile encompassing specific moral values, demographic characteristics, and socioeconomic attributes. Our approach integrates four key components to achieve this:


\textbf{Local Contextual Information.} We ground our synthetic data by incorporating local context drawn from news websites within the target communities. These sources provide relevant contextual information and do not necessarily need to be in the target local language. 


\textbf{Core Cultural Heritage Concepts.} We integrate key local cultural elements, such as cuisine, landmarks, and celebrations, by extracting relevant articles from country-specific Wikipedia portals.


\textbf{Linguistic and Cultural Expressions.} To authentically capture local idiomatic styles, we collect common expressions, proverbs, idioms, dialogues from TV programs, and local terminology, pairing each with English translations for accuracy.

\begin{figure}[htp]
    \centering
    \includegraphics[width=0.45\textwidth]{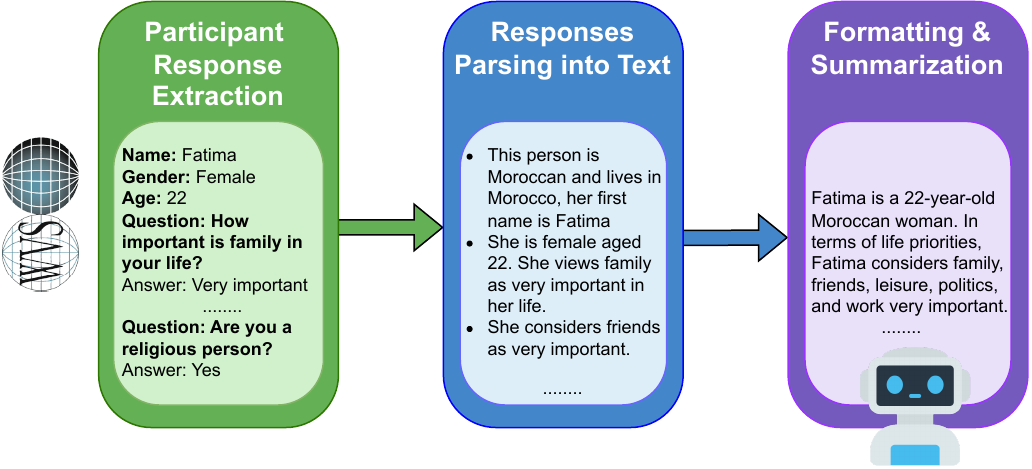} 
    \caption{Pipeline for generation of persona descriptions using the WVS.}
    \label{fig:personas_generation} 
\end{figure}

\textbf{Representative Personas.} We develop representative personas reflecting local moral, demographic, and socioeconomic attributes by leveraging data from the World Values Survey (WVS)~\cite{ZA7505}. Selected survey responses are transformed into textual descriptions, which are further refined by an LLM to create concise and coherent persona profiles (see Figure \ref{fig:personas_generation}).



To produce diverse text genres for pre-training, we combine data points from the four listed components into a unified prompt to guide the teacher LLM. This prompt instructs the LLM to generate varied text outputs in the target low-resource language, explicitly integrating the selected persona’s values, the specified cultural concepts, and provided linguistic cues. Specifically, we focus on generating the following genres: \textit{stories}, \textit{personal essays}, \textit{blog posts}, \textit{reviews}, and \textit{conversations}. An example of this process is depicted in Figure \ref{fig:synthetic_pre_training} (b).

\subsubsection{Retrieval for Local Cultural Heritage} \label{sec:retrieval_pt_data}

This method involves querying a search engine using a pre-defined list of cultural concepts that span multiple cultural categories. For each concept, we extract the top $20$ search results, systematically excluding social media platforms.
The textual content from the retrieved web pages is then parsed and extracted using Trafilatura~\cite{barbaresi-2021-trafilatura}.

\subsection{Model Training}
We evaluate our proposed method on Egyptian (EGY) and Moroccan (MOR) Arabic dialects. Despite their large speaker populations, these dialects remain low-resource, underscoring the need for specialized language models. We select Command R+ \cite{cohere_for_ai_2024} (104B) as our teacher model, as it demonstrates reasonable text-generation capabilities in both target dialects. Additionally, Command R+ provides open weights, enabling us to efficiently generate or translate extensive datasets without incurring API costs.


\subsubsection{Continued Pre-Training}

\paragraph{Data.}
We generate pre-training data for EGY and MOR using the methods outlined in Section~\ref{sec:data_aug}. Our approach involves three main components: \textbf{(i) MT Data.}
We employ our teacher model to translate English educational content into both dialects. Specifically, we sample 5.5 million texts from the Fineweb-edu dataset~\cite{penedo2024finewebdatasetsdecantingweb} and translate them into EGY and MOR. \textbf{(ii) Controlled Synthetic Generation Data.}
We craft tailored prompts incorporating personas, local cultural contexts, dialectal glosses, expressions, and utterances to instruct the LLM in generating diverse genres of text. For \textit{persona} descriptions, we generate 1,200 descriptions based on data from Egyptian and Moroccan participants in the WVS.
For \textit{local news context}, we leverage an in-house corpus comprising approximately 1.5 million Egyptian and 800,000 Moroccan news articles, originally published in MSA by local news websites. Additionally, we include 25,000 Egyptian and 49,000 Moroccan Wikipedia articles. For \textit{dialectal glosses, expressions, and utterances}, we draw from publicly available resources on EGY and MOR proverbs and idiomatic expressions, each accompanied by English explanations. We further augment this with an in-house dataset of 600 dialectal utterances from Egyptian and Moroccan television shows paired with English translations, as well as 4,000 dialect-to-English word pairs for each dialect from the Gatitos dictionary~\cite{jones-etal-2023-gatitos}. \textbf{(iii) Retrieval Data.} For information retrieval, we query the Brave Search API\footnote{\url{https://brave.com/search/api/}} using 6,500 cultural concepts from Morocco and 4,500 cultural concepts from Egypt. These concepts represent the ten cultural heritage categories in the set \textit{\{food, clothes, landmarks, festivals \& celebrations, geography, handicrafts, architecture, fauna, flora, music\}}.

The generated dataset comprises approximately 5.5 million educational articles for both EGY and MOR. Additionally, for EGY, it includes approximately 300,000 samples for each category of conversations, personal essays, blog posts, reviews, and stories. For MOR, there are approximately 150,000 samples for each of these same categories. These latter categories represent the \textit{LHV} dimensions (\S\ref{sec:data_aug}). Table \ref{tab:cultural_tg_examples} presents a sample of these texts.

A filtering process using a repetitive n-gram filter removed 3.97\% of the data. We also conducted a dialectness check on the generated data using ALDi~\cite{keleg-etal-2023-aldi}. The average dialectness scores for the EGY and MOR educational articles are $0.45$ and $0.32$, respectively. In contrast, for the texts focused on cultural heritage and values, the average dialectness scores are higher, at $0.84$ for EGY and $0.72$ for MOR. We attribute the lower dialectness levels in the educational articles to the prevalence of scientific terms that often lack direct equivalents in EGY and MOR, and were therefore retained in MSA. We convert 1.5M EGY and 0.5M MOR samples from the generated data to Arabizi. For retrieval, we collect 110,000 and 30,000 articles about cultural heritage for both EGY and MOR.

Our final pre-training dataset is a mixture of our generated and retrieved data, combined with pre-existing publicly available data for these dialects, MSA, English, French, Math, and Code. Our objective is to preserve the data distribution of the base model's pre-training data to mitigate catastrophic forgetting \cite{luo2025empiricalstudycatastrophicforgetting}. The resulting pre-training dataset comprises 98.57 billion words, and its composition is detailed in Table \ref{tab:data_stats}.

\paragraph{Compute.} We used a cluster of 4$\times$A100 80GB GPUs for 1,096 hours to create our augmented pre-training dataset using the listed inputs.

\paragraph{Continued Pre-training.} Rather than pretraining an LLM from scratch, we continue pretraining Qwen-2.5-3B \cite{qwen2025qwen25technicalreport} with our data. 
We select this model due to its competitive performance and good tokenizer compression ratio on MSA. We continue pretraining the full model (3.1B parameters) for one whole epoch, which took 750 hours on 4$\times$A100 80GB GPUs. More details about the base model selection and the training are in Appendix \ref{app:pre-training}.

\subsubsection{Supervised Fine-Tuning}
To adapt our pre-trained model for instruction following, we perform supervised fine-tuning (SFT).

\paragraph{Data.} Due to the scarcity of SFT datasets for EGY and MOR, we construct a comprehensive training set. This process involves several key steps: (i) translation of SmolTalk dataset \cite{allal2025smollm2smolgoesbig} into MOR, EGY, French, and MSA using the teacher LLM; (ii) synthetic generation of dialectal question-answer pairs using our retrieved dataset of local Egyptian and Moroccan cultural heritage;\footnote{This data is initially created in Arabic script, and a portion is subsequently converted to Arabizi.} (iii) incorporation of the Darija-SFT-Mixture MOR dataset provided by \citet{shang-etal-2025-atlas}; and (iv) translation of TULU-V2-mix dataset \cite{ivison2023camelschangingclimateenhancing} into EGY. Finally, (v) this consolidated SFT dataset is augmented by converting understanding and generation tasks from the training sets of the ORCA \cite{elmadany-etal-2023-orca} and Dolphin \cite{nagoudi-etal-2023-dolphin} benchmarks into instruction-response formats. The final composition of our instruction dataset is in Table \ref{tab:inst_data_distro}.

\paragraph{Fine-Tuning.} For model SFT, we follow recent approaches \cite{ramé2024warpbenefitsweightaveraged, dang2024ayaexpansecombiningresearch} that leverage model merging techniques to produce models effective across multiple languages or tailored for particular tasks. Specifically, we fully fine-tuned two separate variants of the base model—one specialized for MOR and the other for EGY—each trained on its respective dialectal data in both Arabic script and Arabizi (plus an amount of shared data between the two variants from the other languages; see~\ref{app:sft_phase}). We fine-tune each dialect-specific model for two epochs and employ weighted linear averaging~\cite{aakanksha2024mixdatamergemodels} for merging, dubbing our merged model \textit{NileChat}. 

More information about our model merging is in~\ref{app:sft_phase}  and the prompts used for generating and translating our pre- and fine-tuning datasets is in~\ref{app:prompts}.





\section{Experiments}
\subsection{Evaluation Tasks}
We employ a comprehensive evaluation framework to measure the performance of \textit{NileChat} for EGY and MOR. This framework enables comparison with our baseline and other LLMs across multiple capability dimensions: Understanding, cultural knowledge, translation, and value alignment.

\paragraph{Understanding.}
We evaluate understanding capabilities using MMLU~\cite{hendrycks2021measuring}, HellaSwag~\cite{zellers-etal-2019-hellaswag}, and Belebele~\cite{bandarkar-etal-2024-belebele} benchmarks, each adapted to both EGY and MOR dialects. For MOR, we directly employ the MMLU and HellaSwag versions provided by~\citet{shang-etal-2025-atlas}. For EGY, we follow the translation pipeline described in~\citet{shang-etal-2025-atlas}, translating the English and MSA MMLU tasks and the English HellaSwag dataset into EGY using our teacher model~\footnote{We have publicly released the EgyMMLU and EgyHellaswag benchmarks for evaluation on the LM Evaluation Harness Framework at \url{https://github.com/EleutherAI/lm-evaluation-harness/tree/main/lm_eval/tasks/egymmlu} and \url{https://github.com/EleutherAI/lm-evaluation-harness/tree/main/lm_eval/tasks/egyhellaswag}.}.
A careful verification of the translation quality for the generated EGY MMLU and EGY HellaSwag shows that the average correctness is 3.85 on a scale from 1-5 and the average dialectness is approximately 4 on a scale from 1-5. Further details are provided in Appendix~\ref{app:tasks}.
For the Belebele benchmark, we utilize the official Moroccan and Egyptian dialect sets. Evaluations are conducted in both zero-shot and 3-shot scenarios, using accuracy as our performance metric.

\paragraph{Cultural Knowledge.}
To assess cultural knowledge specific to Morocco and Egypt, we utilize the publicly available test set from the Palm benchmark~\cite{alwajih2025palmculturallyinclusivelinguistically}, focusing on these two countries only. We adopt an LLM-as-Judge methodology~\cite{zheng2023judging}, employing Gemma-3-27b~\cite{gemmateam2025gemma3technicalreport} to rate the correctness of model-generated responses compared to ground-truth answers on a scale from 0 to 10. The final evaluation score is calculated as the average correctness across all responses.

\paragraph{Translation.}
We evaluate the translation performance across multiple directions: \textit{dialect$\leftrightarrow$dialect} (i.e., Moroccan$\leftrightarrow$Egyptian), \textit{dialect$\leftrightarrow$MSA}, \textit{English$\leftrightarrow$dialect}, and \textit{French$\leftrightarrow$dialect}. Our primary benchmark is the Flores-200 dataset~\cite{nllbteam2022languageleftbehindscaling}, comprising $1,012$ test examples per translation direction.
Additionally, we introduce an in-house, human-curated dataset consisting of 300 authentic EGY and MOR utterances transcribed from local television programs then translated to MSA and English. This dataset provides a more accurate reflection of natural, colloquial language usage compared to Flores-200, which primarily contains Wikipedia-based sentences. We conduct evaluations in both zero-shot and 4-shot settings, reporting results using ChrF++~\cite{popovic-2015-chrf} and spBLEU scores~\cite{goyal-etal-2022-flores}.


\paragraph{Value Alignment.}
To assess alignment with societal values, we adapt WVS questions into a multiple-choice format (expressed in the local language). The questions are categorized into 13 dimensions such as \textit{Economic Values (EcoV)}, \textit{Ethical Values (EthV)}, and \textit{Happiness and Wellbeing (HW)}.\footnote{See Appendix \ref{app:tasks} for the full list.} We use the Social Value Alignment (SVA) metric \citep{lee-etal-2024-kornat}, which measures alignment using the distribution of survey responses. A model's alignment score for each question corresponds to the proportion of participants who chose the model-predicted option, averaged across all questions for the final score.

\paragraph{Baseline Models.}
We compare \textit{NileChat} against a set of $17$ instruction-tuned LLMs known for their strong capabilities in Arabic, capped at 13B parameters (see full list in Table \ref{tab:models_params} and Appendix \ref{app:baselines} for details).\footnote{We also evaluate our translation performance against an NLLB-200's 3.3B variant \cite{nllbteam2022languageleftbehindscaling}.}




\subsection{Results and Discussion} \label{sec:results}

\begin{table*}[]
\centering
\resizebox{.90\textwidth}{!}{%
\begin{tabular}{ll|cc|cc|cc|cc}
\toprule
\multicolumn{2}{c}{\multirow{2}{*}{\textbf{Model}}} & \multicolumn{2}{c|}{\textbf{MMLU}} & \multicolumn{2}{c|}{\textbf{HellaSwag}} & \multicolumn{2}{c|}{\textbf{Belebele}} & \multicolumn{2}{c}{\textbf{Palm}} \\
\cmidrule(lr){3-4} \cmidrule(lr){5-6} \cmidrule(lr){7-8} \cmidrule(lr){9-10}
\multicolumn{2}{c}{} & \multicolumn{1}{l}{\textbf{EGY}} & \multicolumn{1}{l|}{\textbf{MOR}} & \multicolumn{1}{l}{\textbf{EGY}} & \multicolumn{1}{l|}{\textbf{MOR}} & \multicolumn{1}{l}{\textbf{EGY}} & \multicolumn{1}{l|}{\textbf{MOR}} & \multicolumn{1}{l}{\textbf{EGY}} & \multicolumn{1}{l}{\textbf{MOR}} \\
\midrule
\multirow{8}{*}{\rotatebox{90}{\textbf{Less than 7B}}} & \textbf{Qwen3-1.7B} & 28.53 & 28.53 & 28.44 & 27.47 & 22.89 & 22.89 & 3.61 & 2.12 \\
& \textbf{ar-stablelm-2-chat} & 41.56 & 40.36 & 34.79 & 33.45 & 38.89 & 36.11 & 4.20 & 3.62 \\
& \textbf{Atlas-Chat-2B} & 42.61 & 44.87 & 29.66 & 34.74 & 50.56 & 55.67 & 3.16 & 3.42 \\
& \textbf{Llama-3.2-3B-Instruct} & 40.68 & 37.54 & 29.16 & 28.27 & 45.44 & 35.89 & 3.21 & 2.28 \\
& \textbf{gemma-3-4b-it} & 40.79 & 32.70 & 34.21 & 31.35 & 37.33 & 34.22 & \textbf{7.61} & 5.42 \\
& \textbf{Qwen3-4B} & 28.61 & 28.54 & 30.28 & 29.04 & 22.89 & 22.89 & 4.51 & 2.71 \\
& \textbf{Qwen2.5-3B-Instruct} & 43.37 & 44.43 & 31.62 & 29.58 & 51.33 & 41.44 & 2.86 & 2.31 \\
& \textbf{\colorbox{mylightyellow}{NileChat (3B)}} & \textbf{57.56} & \textbf{57.36} & \textbf{37.97} & \textbf{39.33} & \textbf{72.67} & \textbf{70.33} & 5.72 & \textbf{5.86} \\
\midrule
\multirow{8}{*}{\rotatebox{90}{\textbf{More than 7B}}} & \textbf{AceGPT-7B-chat} & 40.29 & 37.57 & 33.27 & 30.47 & 32.67 & 32.00 & 5.58 & 3.93 \\
& \textbf{ALLaM-7B-Instruct} & 60.04 & 58.72 & \underline{39.40} & 37.30 & 69.56 & 57.78 & 6.78 & 6.14 \\
& \textbf{Qwen2.5-7B-Instruct} & 49.65 & 44.98 & 34.67 & 32.16 & 64.22 & 48.56 & 6.70 & 4.77 \\
& \textbf{Qwen3-8B} & 28.53 & 28.53 & 31.76 & 30.32 & 22.89 & 22.89 & 5.88 & 3.96 \\
& \textbf{Atlas-Chat-9B} & 55.17 & 58.84 & 33.71 & \underline{44.34} & 70.33 & \underline{74.11} & 5.24 & 4.84 \\
& \textbf{gemma-3-12b-it} & \underline{61.17} & \underline{60.00} & 38.59 & 35.66 & \underline{75.78} & 64.89 & \underline{8.76} & \underline{7.09} \\
& \textbf{AceGPT-13B-chat} & 45.45 & 40.68 & 35.06 & 32.40 & 38.78 & 36.44 & 6.10 & 4.83 \\
& \textbf{jais-13b-chat} & 49.79 & 48.10 & 39.02 & 36.56 & 64.22 & 53.78 & 5.66 & 4.80 \\
\bottomrule
\end{tabular}%
}
\caption{Zero-shot performance of models on understanding and cultural knowledge evaluations. Metrics are accuracy for MMLU, HellaSwag, and Belebele, and a 0-10 correctness score for Palm. Bold values indicate the highest score among models comparable in size to ours (< 7B). Underlined values represent the highest score in the entire column, including larger models.}
\label{tab:unders_cult_results}
\end{table*}

\paragraph{Understanding.} 
As Table \ref{tab:unders_cult_results} shows, \textit{NileChat} demonstrates SoTA performance on the MMLU, HellaSwag, and Belebele benchmarks for both EGY and MOR when compared to similar size models. Specifically, \textit{NileChat} surpasses its baseline model, Qwen2.5-3B-instruct, by a significant margin of $\sim$10 points across the majority of these tasks. Notably, \textit{NileChat} also outperforms larger Arabic-focused models such as AceGPT-13B and Jais-13B. Furthermore, it achieves on-par performance with recent leading Arabic LLMs like ALLaM-7B, with a performance gap of less than 3 points on most tasks, and even surpasses it on certain benchmarks, including Belebele. Results for 3-shot are presented in Table \ref{tab:three_shot_unders} and they show a similar trend to the zero-shot ones.

\paragraph{Cultural Knowledge.} 
As shown in Table~\ref{tab:unders_cult_results}, our approach significantly enhances cultural knowledge (Palm), enabling \textit{NileChat} to achieve scores of 5.72 (EGY) and 5.86 (MOR), compared to baseline Qwen2.5-3B-instruct scores of 2.86 and 2.31, respectively. Among similarly sized models, ours achieves the highest performance on MOR and ranks second only to Gemma-3-4B for EGY. Although larger models such as Gemma-3-12B exhibit superior overall scores (EGY: 8.71, MOR: 7.09), \textit{NileChat} notably surpasses AceGPT-7B and -13B on Moroccan cultural knowledge, despite their claimed alignment with Arabic cultures. Additionally, it outperforms Atlas-chat-2B and -9B, models specifically fine-tuned for Moroccan dialects. \textit{These results support our claim that linguistic fluency alone—gained through supervised fine-tuning or pre-training on potentially biased, translated datasets—is insufficient for genuine cultural alignment with local communities.}

\paragraph{Translation.} Table~\ref{tab:translation_evals_agg} summarizes the spBLEU scores from our zero-shot translation. Overall, \textit{NileChat} achieves the highest average translation quality (spBLEU: 21.32), outperforming all evaluated models, including larger alternatives such as ALLaM-7B (20.60) and NLLB-200-3.3B (18.29). Specifically, on the Flores benchmark, \textit{NileChat} demonstrates comparable performance to the similarly-sized NLLB-200-3.3B, with only a marginal 1-point spBLEU difference aggregated across MOR and EGY. Notably, \textit{NileChat} surpasses even larger competitors in all translation directions, except when translating into MOR, where its performance matches that of Atlas-Chat-9B—a larger, single-dialect-focused model that is 3X larger.

On our in-house, human-curated dataset—which closely represents authentic speech patterns from local populations—\textit{NileChat} significantly outperforms all baselines, including NLLB-200-3.3B, in all translation directions for both EGY and MOR. This real-world evaluation emphasizes the effectiveness of our strategy to incorporate local linguistic and cultural elements into synthetic data generation, enriching the pre-training data with diverse dialectal expressions and vocabulary. Detailed results for both zero-shot and 4-shot translation experiments are provided in Table~\ref{tab:combined_flores_inhouse_stacked}.

\begin{table*}[!ht]
\centering
\resizebox{.90\textwidth}{!}{
\begin{tabular}{llcccc|cccc|c}
\toprule
& \multirow{2}{*}{\textbf{Model}} & \multicolumn{4}{c}{\textbf{Flores-200}} & \multicolumn{4}{c}{\textbf{In-House Data}} & \multicolumn{1}{c}{\multirow{2}{*}{\textbf{Average}}} \\ \cmidrule(lr){3-6} \cmidrule(lr){7-10}
& & \multicolumn{2}{c}{\textbf{XX $\rightarrow$}} & \multicolumn{2}{c|}{\textbf{ $\rightarrow$ XX}} & \multicolumn{2}{c}{\textbf{XX $\rightarrow$}} & \multicolumn{2}{c|}{\textbf{$\rightarrow$ XX}} & \multicolumn{1}{c}{} \\ \cmidrule(lr){3-4} \cmidrule(lr){5-6} \cmidrule(lr){7-8} \cmidrule(lr){9-10}
& & \multicolumn{1}{c}{\textbf{EGY}} & \multicolumn{1}{c}{\textbf{MOR}} & \multicolumn{1}{c}{\textbf{EGY}} & \multicolumn{1}{c|}{\textbf{MOR}} & \multicolumn{1}{c}{\textbf{EGY}} & \multicolumn{1}{c}{\textbf{MOR}} & \multicolumn{1}{c}{\textbf{EGY}} & \multicolumn{1}{c|}{\textbf{MOR}} & \multicolumn{1}{c}{} \\ \midrule
\multirow{9}{*}{\rotatebox[origin=c]{90}{\textbf{Less than 7B}}} & \textbf{Qwen3-1.7B} & 14.75 & 10.89 & 19.51 & 15.47 & 11.41 & 4.36 & 15.63 & 6.32 & 12.29 \\
& \textbf{ar-stablelm-2-chat} & 14.35 & 7.07 & 11.10 & 9.72 & 9.23 & 2.92 & 11.23 & 7.73 & 9.17 \\
& \textbf{Atlas-Chat-2B} & 15.20 & 13.40 & 21.39 & 21.11 & 5.36 & 7.83 & 14.52 & 13.54 & 14.05 \\
& \textbf{Llama-3.2-3B-Instruct} & 14.25 & 9.15 & 19.28 & 15.54 & 10.67 & 3.16 & 13.61 & 4.87 & 11.32 \\
& \textbf{gemma-3-4b-it} & 9.27 & 5.22 & 12.46 & 10.13 & 3.01 & 0.60 & 16.89 & 5.25 & 7.86 \\
& \textbf{Qwen3-4B} & 17.93 & 11.64 & 20.03 & 18.90 & 13.09 & 4.44 & 20.72 & 8.52 & 14.41 \\
& \textbf{NLLB-200-3.3B} & \underline{\textbf{23.93}} & 15.37 & \underline{\textbf{25.84}} & \underline{\textbf{26.57}} & 16.77 & 7.49 & 18.90 & 11.43 & 18.29 \\
& \textbf{Qwen2.5-3B-Instruct} & 15.14 & 11.27 & 20.52 & 17.37 & 9.91 & 4.19 & 19.24 & 7.83 & 13.18 \\
& \textbf{\colorbox{mylightyellow}{NileChat (3B)}} & 23.60 & \textbf{16.41} & 25.74 & 25.56 & \underline{\textbf{22.02}} & \underline{\textbf{12.34}} & \textbf{26.50} & \textbf{18.39} & \underline{\textbf{21.32}} \\ \midrule
\multirow{8}{*}{\rotatebox[origin=c]{90}{\textbf{More than 7B}}} & \textbf{AceGPT-7B-chat} & 18.02 & 11.33 & 21.11 & 17.46 & 14.73 & 4.95 & 20.10 & 7.47 & 14.40 \\
& \textbf{ALLaM-7B-Instruct} & 23.91 & 15.88 & 24.74 & 23.19 & 19.98 & 9.16 & \underline{29.40} & \underline{18.51} & 20.60 \\
& \textbf{Qwen2.5-7B-Instruct} & 14.41 & 10.23 & 19.81 & 18.95 & 10.43 & 4.10 & 20.92 & 8.80 & 13.46 \\
& \textbf{Qwen3-8B} & 20.03 & 13.86 & 22.56 & 21.33 & 13.38 & 4.73 & 24.14 & 9.27 & 16.16 \\
& \textbf{Atlas-Chat-9B} & 18.20 & \underline{16.89} & 24.92 & 26.29 & 5.36 & 7.68 & 17.35 & 15.23 & 16.49 \\
& \textbf{gemma-3-12b-it} & 13.01 & 4.89 & 19.05 & 19.54 & 7.86 & 2.45 & 24.51 & 12.38 & 12.96 \\
& \textbf{AceGPT-13B-chat} & 19.48 & 14.02 & 22.81 & 19.84 & 15.54 & 5.56 & 23.51 & 9.52 & 16.29 \\
& \textbf{jais-13b-chat} & 8.80 & 4.29 & 15.77 & 17.12 & 10.83 & 4.02 & 19.19 & 12.47 & 11.56 \\ \bottomrule
\end{tabular}}
\caption{Zero-shot translation performance (spBLEU) on the Flores and in-house datasets. XX $\rightarrow$ EGY and XX $\rightarrow$ MOR denote average over target languages EGY and MOR, respectively. Conversely, EGY → XX and MOR → XX indicate average over EGY and MOR as source languages. Bold values highlight the top score among models with fewer than 7 billion parameters. Underlined values indicate the highest score overall in each column. Detailed results are in Table \ref{tab:combined_flores_inhouse_stacked}.}
\label{tab:translation_evals_agg}
\end{table*}

\begin{figure*}[!ht]
    \centering
    \includegraphics[width=0.99\textwidth, trim={0pt 0pt 130pt 0pt}, clip]{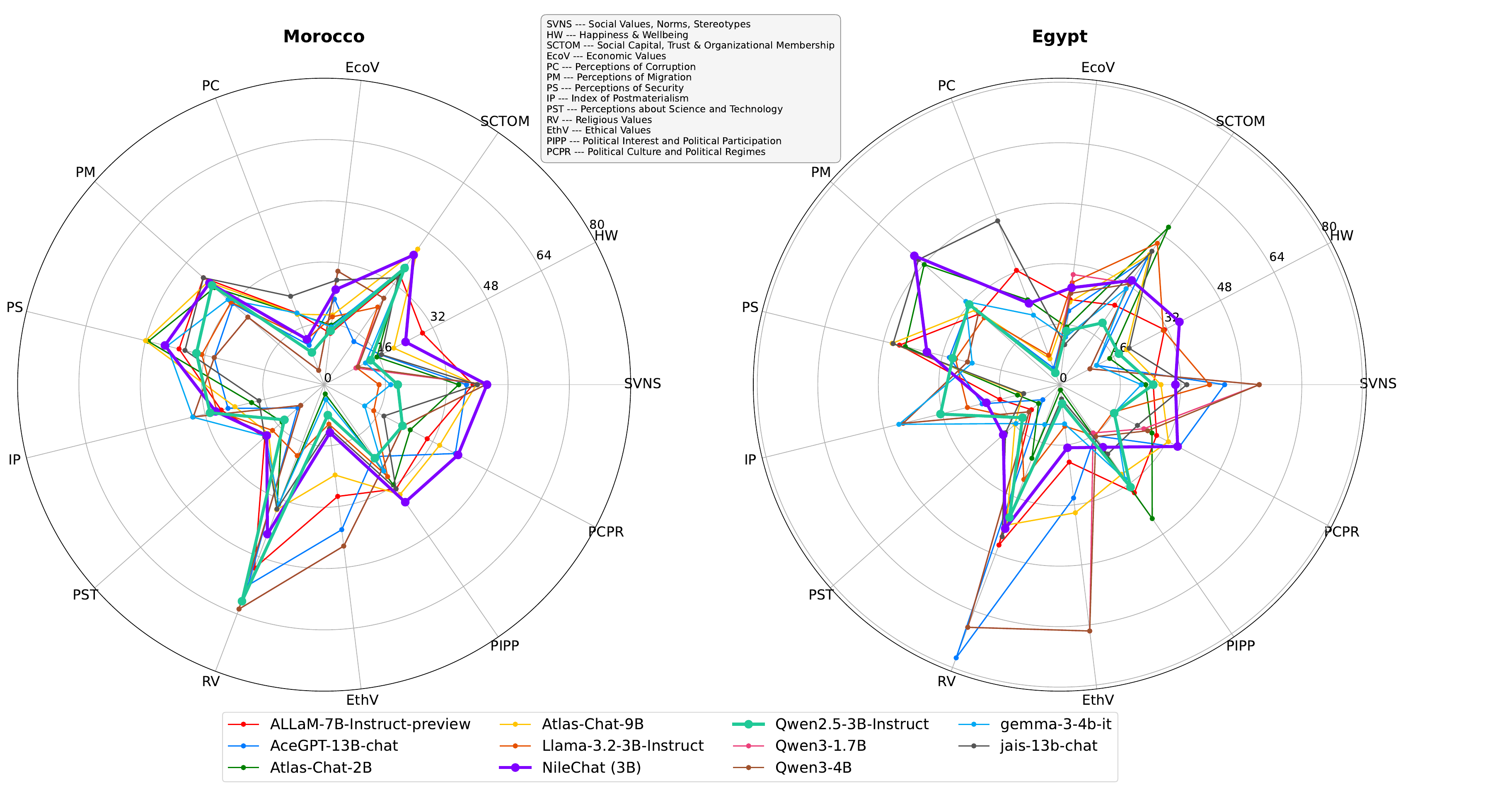} 
    \caption{Average SVA scores of evaluated models across societal value dimensions for Egypt and Morocco.}
    \label{fig:wvs_results} 
\end{figure*}

\paragraph{Value Alignment.} Figure~\ref{fig:wvs_results} illustrates the results of value alignment evaluation based on the WVS. \textit{NileChat} demonstrates substantial improvements over the baseline across most societal-value dimensions for both Moroccan and Egyptian contexts. Specifically, for Morocco, \textit{NileChat} surpasses the baseline in all dimensions except Religious Values and the Index of Postmaterialism. Similarly, for Egypt, it outperforms the baseline across all dimensions except Political Interest and Political Participation, and the Index of Postmaterialism. These findings indicate that our approach—where a teacher LLM engages in role-playing by generating diverse text genres through personas embodying local community values—successfully steers responses towards culturally aligned positions. In a broader comparative analysis against all evaluated models, ours achieves the best results for Morocco across several dimensions and remains competitive in others. For Egypt, \textit{NileChat} notably excels in Perceptions of Migration, Political Culture and Political Regimes, Happiness and Wellbeing, and Perceptions about Science and Technology, though models such as Jais-13B and ALLaM-7B show slightly stronger performance in certain other dimensions.

\begin{figure}[!ht]
    \centering
    \includegraphics[width=0.48\textwidth]{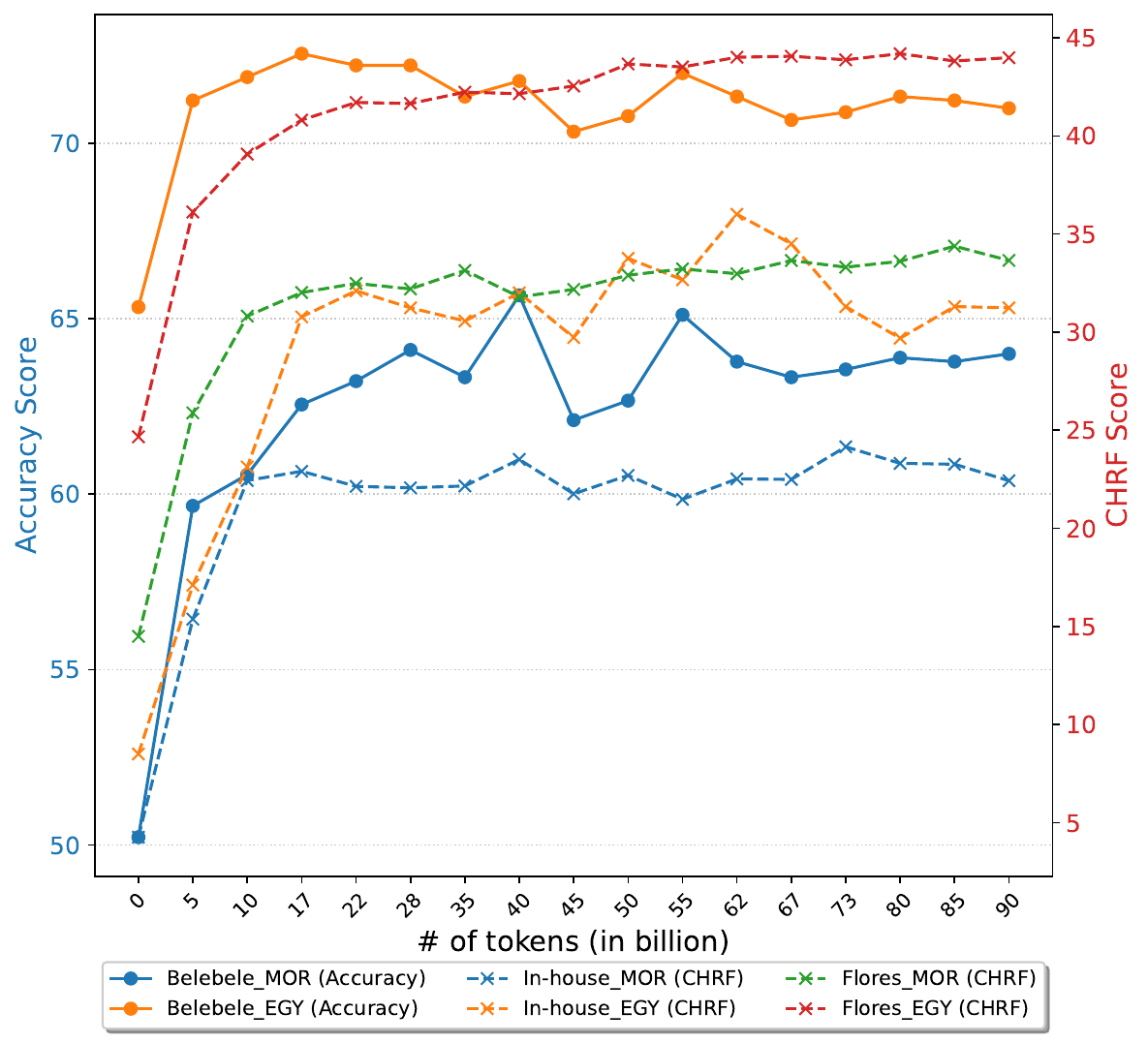} 
    \caption{Evolution of model performance during pre-training, measured by the number of tokens processed.
    }
    \label{fig:tokens_evol} 
\end{figure}

\paragraph{How many pre-training tokens are needed to reach good performance for a new language?} Figure \ref{fig:tokens_evol} shows the performance evolution of \textit{NileChat} during the pre-training phase on Belebele and translation tasks. The charts show that the model starts to get a large boost in these tasks during the first 10B tokens and then continues to slightly increase until it becomes steady after around 60B tokens.

\section{Conclusion}

We introduced a novel methodology for adapting LLMs to specific communities by deeply integrating their unique linguistic characteristics, cultural heritage, and societal values. Our approach leverages a teacher model proficient in generating low-resource languages to enable: (i) translation for the incorporation of community-specific language and (ii) controlled generation and retrieval mechanisms for the authentic inclusion of cultural heritage and values. We validated our methodology using the Moroccan and Egyptian Arabic dialects as testbeds by developing \textit{NileChat}, an LLM covering these two dialects. Comprehensive evaluations on understanding, translation, and cultural alignment benchmarks demonstrate that our method significantly enhances the baseline LLM's performance in capturing target language nuances and cultural values. Notably, \textit{NileChat} also outperforms existing Arabic-aware LLMs. Our method offers a promising research direction for fostering inclusivity of diverse local communities within LLM development, thereby emphasizing the critical role of such an inclusion in the broader democratization of this technology.

\section*{Limitations}


\begin{itemize}
    \item \textbf{Teacher Model Dependency for Low-Resource Languages:} Our method's reliance on a teacher model proficient in generating even low-resource target languages may not hold for extremely under-resourced languages (e.g., Berber, Malayo-Polynesian varieties) \cite{nllbteam2022languageleftbehindscaling}, potentially limiting its applicability in such contexts.
    \item \textbf{Supervised Fine-Tuning Data:} SFT phase predominantly utilized translated data due to resource constraints. This reliance on translated, rather than native, data for SFT might impact the model's nuanced performance in the target languages and their corresponding cultures.
    \item \textbf{Susceptibility to Hallucination:} As a 3B parameter model, our LLM is relatively small, rendering it more prone to hallucination and the generation of inaccurate or incomplete information compared to larger architectures \cite{wei2022emergent}.
    \item \textbf{Computational Cost of Synthetic Data Generation:} The process of generating synthetic data is computationally intensive, particularly when employing large teacher models (e.g., Command R+, a 104B parameter model requiring substantial GPU resources: 4x80GB). This challenge is amplified by the autoregressive generation of long documents from extensive input contexts (e.g., articles, persona descriptions, cultural concepts) restricting the scale of this approach for more languages.
    \item \textbf{Absence of Explicit Safety Alignment:} The model has not undergone dedicated safety alignment. While trained on curated datasets (Wikipedia, educational, news) largely devoid of toxic content and leveraging a safety-aligned teacher LLM, specific safety tuning is acknowledged as important future work.
    \item \textbf{Limited generation of subtle details.} While the controlled generation uses multiple sources (WVS, news, Wikipedia, TV scripts), the generated texts are limited in terms of the very subtle cultural nuances, implicit knowledge, humor, or sarcasm that are often not explicitly stated in these source materials (Wikipedia and news articles).

\end{itemize}

\section*{Ethics Statement}

Our work contributes to the development of inclusive, linguistically, and culturally diverse LLMs capable of serving varied communities. While we generate our pre-training and instruction-tuning data using a teacher LLM, this process is critically informed by ground-truth cultural values survey data from the communities of interest and local context to control the generation. This approach aims to imbue our models with specific cultural nuances relevant to these communities.

As our evaluations demonstrate, the resulting models exhibit reasonable alignment with the cultural heritage and values of our target communities and can produce fluent text in their respective dialects. Despite these advancements, we have not conducted explicit safety alignment procedures for these models. Consequently, we strongly recommend thorough testing and further safety evaluations before any deployment in real-world scenarios.

\section*{Acknowledgments}\label{sec:acknow}
Muhammad Abdul-Mageed acknowledges support from Canada Research Chairs (CRC), the Natural Sciences and Engineering Research Council of Canada (NSERC; RGPIN-2018-04267), the Social Sciences and Humanities Research Council of Canada (SSHRC; 895-2020-1004; 895-2021-1008), Canadian Foundation for Innovation (CFI; 37771), Digital Research Alliance of Canada,\footnote{\href{https://alliancecan.ca}{https://alliancecan.ca}} and UBC Advanced Research Computing-Sockeye.\footnote{\href{https://arc.ubc.ca/ubc-arc-sockeye}{https://arc.ubc.ca/ubc-arc-sockeye}}

\bibliography{anthology,custom}

\appendix
\onecolumn
\counterwithout{table}{section}

\counterwithin{table}{section}

\renewcommand{\thetable}{\Alph{section}.\arabic{table}}

\section{Data} \label{app:data}

\begin{table}[]
\centering
\resizebox{.9\textwidth}{!}{%
\begin{tabular}{llllll}
\toprule
\textbf{Language}                               & \multicolumn{2}{l}{\textbf{Data category}}                                  & \textbf{Data source}                                                                        & \textbf{Nature}    & \textbf{\# of words} \\
\midrule
\multirow{10}{*}{\textbf{Dialectal data}}      & \multirow{5}{*}{\textbf{Egypt}}           & \multirow{4}{*}{General}    & Wikipedia                                                                                   & Real               & 128.71M                  \\
                                                &                                           &                             & \textbf{MT fineweb-EDU}                                                                     & \textbf{Synthetic} & \textbf{2.08B}           \\
                                                &                                           &                             & \textbf{LHV}                                                                              & \textbf{Synthetic} & \textbf{398.89M}         \\
                                                &                                           &                             & Fineweb2                                                                                    & Real               & 430.46M                  \\
                                                \cmidrule(l){3-6} 
                                                &                                           & Arabizi                     & \textbf{MT fineweb-EDU (Arabizi) \& LHV}                                                  & \textbf{Synthetic} & \textbf{206.49M}         \\
                                                \cmidrule(l){2-6} 
                                                & \multirow{5}{*}{\textbf{Morocco}}         & \multirow{4}{*}{General}    & Wikipedia                                                                                   & Real               & 1.67M                    \\
                                                &                                           &                             & \textbf{Translated fineweb-EDU}                                                             & \textbf{Synthetic} & \textbf{2.02B}           \\
                                                &                                           &                             & \textbf{LHV}                                                                              & \textbf{Synthetic} & \textbf{207.41M}         \\
                                                &                                           &                             & Fineweb2                                                                                    & Real               & 1.64B                    \\
                                                \cmidrule(l){3-6} 
                                                &                                           & Arabizi                     & \textbf{MT fineweb-EDU (Arabizi) \& LHV}                                                  & \textbf{Synthetic} & \textbf{467.30M}         \\
\midrule
\multirow{6}{*}{\textbf{MSA}}                   & \multirow{2}{*}{\textbf{Egypt}}           & Cultural                    & \textbf{Brave API}                                                                          & \textbf{Real}      & \textbf{74.67M}          \\
                                                &                                           & General                     & Local News                                                                                  & Real               & 346.79M                  \\
                                                \cmidrule(l){2-6} 
                                                & \multirow{2}{*}{\textbf{Morocco}}         & Cultural                    & \textbf{Brave API}                                                                          & \textbf{Real}      & \textbf{23.08M}          \\
                                                &                                           & General                     & Local News                                                                                  & Real               & 220.16M                  \\
                                                \cmidrule(l){2-6} 
                                                & \multirow{2}{*}{\textbf{General}}         & \multirow{2}{*}{General}    & Fineweb2                                                                                    & Real               & 28.80B                   \\
                                                &                                           &                             & Wikipedia                                                                                   & Real               & 318.62M                  \\
\midrule
\textbf{English}                                & \textbf{General}                          & General                     & Fineweb-EDU                                                                                 & Real               & 51.57B                   \\
\midrule
\textbf{French}                                 & \textbf{General}                          & General                     & Fineweb2                                                                                    & Real               & 9.42B                    \\
\midrule
\textbf{Code \& Math}                           & \textbf{}                                 & Code \& Math                & \begin{tabular}[c]{@{}l@{}}MathGenie/MathCode-Pile\\ macrocosm-os/code-parrot-github-code\end{tabular} & Real               & 818.35M                  \\
\bottomrule
\end{tabular}%
}
\caption{Distribution of the final pre-training data mixture by language, nature (synthetic vs. real), and word count per dataset. Bold rows highlight data generated via our proposed augmentation pipeline.}
\label{tab:data_stats}
\end{table}

\begin{table}[]
\centering

\begin{tabular}{@{} >{\bfseries}l l l r @{}}
\toprule
\textbf{Dataset name} & \textbf{Language} & \textbf{Source} & \textbf{\# of Instructions} \\
\midrule
Darija-SFT-Mixture & MOR (Arabic) & Atlas-Chat & 458,155 \\\midrule
TÜLU-V2-mix & EGY (Arabic) & Ours (MT) & 178,109 \\
\midrule 
\multirow{7}{*}{SmolTalk} & MOR (Arabic) & Ours (MT) & 192,266 \\
 & MOR (Arabizi) & Ours (MT) & 93,419 \\
 & EGY (Arabic) & Ours (MT) & 195,260 \\
 & EGY (Arabizi) & Ours (MT) & 93,181 \\
 & French & Ours (MT) & 99,468 \\
 & MSA & Ours (MT) & 96,933 \\
 & English & SmolTalk & 149,124 \\
\midrule 
ORCA & \multirow{2}{*}{MSA + dialects} & Ours (Converted) & 460,203 \\
Dolphin & & Ours (Converted) & 425,703 \\
\midrule 
\multirow{2}{*}{Cultural instructions} & MOR (Arabic) & Ours (Synthetic) & 25,159 \\
 & EGY (Arabic) & Ours (Synthetic) & 107,428 \\
\bottomrule
\end{tabular}
\caption{Distribution of the final instruction and response data mixture by language and number of instructions per dataset. 'Ours' refers to datasets we created via machine translation (MT) or by converting existing datasets into an instruction/response format.}
\label{tab:inst_data_distro}
\end{table}

\begin{table*}[t]
\centering 
\includegraphics[width=0.95\linewidth, trim={0pt 185pt 0pt 0pt}, clip]{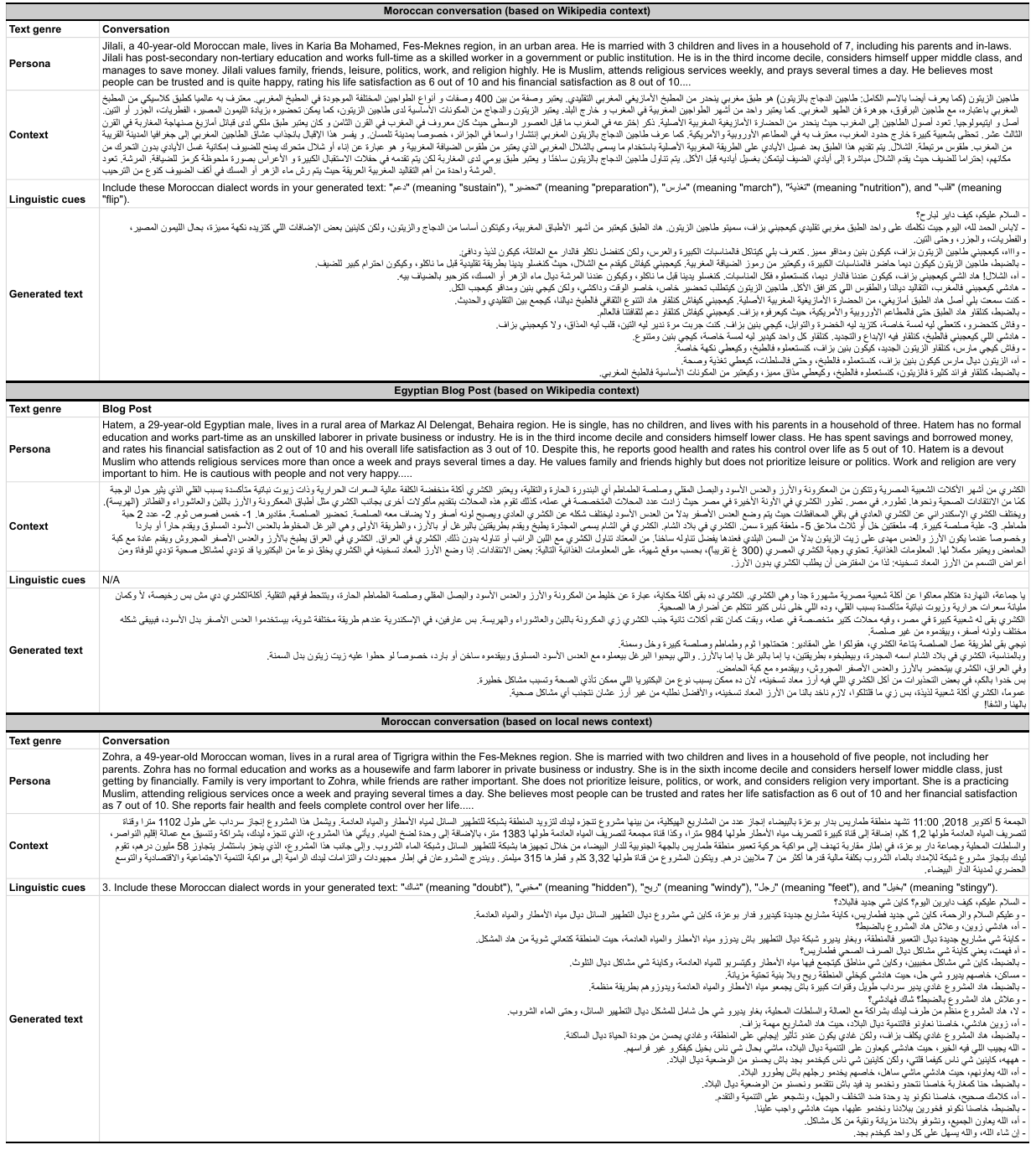}
\caption{Examples of culturally aware and dialectally diverse texts from various genres, generated by our teacher model. The model was provided with input representing a specific persona, local context, and a local linguistic cue following the methodology described in Section \ref{sec:data_aug}.}
\label{tab:cultural_tg_examples}
\end{table*}


\section{Training details}
\subsection{Pre-training} \label{app:pre-training}

\textbf{Motivations for choosing Qwen-2.5-3B as our backbone model.} We select Qwen-2.5-3B as our base model to continue pretrain for two primary reasons: its competitive performance on MSA tasks and good tokenizer compression ratio on Arabic dialect texts. At the time of the selection, the Gemma 2 \cite{gemmateam2024gemma2improvingopen} and Qwen 2.5 base models showed the best performance in MSA. Also, both of their tokenizers have a good compression ratio for Arabic text in both standard and dialectal forms. Our analysis shows a ratio between 2.7 and 2.8 for Gemma, while Qwen 2.5 has a compression ratio between 2.9 and 3.1. Our final choice of Qwen 2.5 was based on its better performance on MSA.

\textbf{Details for continued model pretraining.} We continue the pre-training of Qwen-2.5-3B on our curated pre-training dataset. Subsequently, the model is fully fine-tuned for one epoch using a sequence length of 4,096. To optimize the learning process, the learning rate is linearly decayed from $5 \times 10^{-6}$ to $5 \times 10^{-7}$. To mitigate overfitting, we apply a weight decay of $0.1$, and gradient norms are clipped at a maximum value of $1.0$. The training is performed on a cluster of 4$\times$A100 80GB GPUs.

\subsection{Supervised fine-tuning} \label{app:sft_phase}
To enhance model robustness and facilitate effective merging, we augmented each dialect-specific dataset with a shared multilingual corpus, comprising English SmolTalk, MSA SmolTalk, French SmolTalk, and additional data from the ORCA and Dolphin datasets. Each dialect-specific model was trained for two epochs with a sequence length of 4,096 tokens, using a learning rate that linearly decayed from $7 \times 10^{-6}$ to $7 \times 10^{-7}$. 

Table \ref{tab:merging_diff} compares the SFT model with models fine-tuned on Egyptian and Moroccan datasets individually, as well as with our final merged model, \textit{NileChat}. \textit{NileChat} performs well on tasks for both EGY and MOR. The MOR-specific model also demonstrates strong performance on both MOR and, to some extent, EGY tasks. In contrast,  the EGY-specific model does not perform well on MOR tasks.

Table \ref{tab:merging_diff} compares the SFT model with models fine-tuned on Egyptian and Moroccan datasets individually, as well as with our final merged model, NileChat. NileChat performs well on tasks for both EGY and MOR. The MOR-specific model also demonstrates strong performance on both MOR and, to some extent, EGY tasks. In contrast, the EGY-specific model does not perform well on MOR tasks. We relate this observed asymmetry to the linguistic characteristics of the dialects relative to MSA. During the SFT phase, each dialect-specific dataset was augmented with a shared multilingual corpus which included MSA data (e.g., MSA SmolTalk, and data from the ORCA and Dolphin datasets). It is plausible that EGY is linguistically closer to MSA compared to the MOR, which is often considered more distant from MSA due to influences such as Berber and French. Consequently, the MOR-tuned model, having been exposed to this shared MSA data, might more effectively leverage this MSA knowledge to generalize to EGY tasks. Conversely, the greater linguistic divergence of the Moroccan dialect from MSA could make it more challenging for the EGY-tuned model to transfer its learning, including the MSA component, to the distinct features of the Moroccan dialect.

\begin{table}[]
\centering
\resizebox{.98\textwidth}{!}{%
\begin{tabular}{lcccccc}
\toprule
& \multicolumn{2}{c}{\textbf{Belebele}} & \multicolumn{2}{c}{\textbf{Flores}} & \multicolumn{2}{c}{\textbf{In-house}} \\
\cmidrule(lr){2-3} \cmidrule(lr){4-5} \cmidrule(lr){6-7}
& \textbf{MOR} & \textbf{EGY} & \textbf{ENG$\rightarrow$EGY} & \textbf{ENG$\rightarrow$MOR} & \textbf{ENG$\rightarrow$MOR} & \textbf{ENG$\rightarrow$EGY} \\
\midrule
\textbf{NileChat-EGY} & 64.44 & 70.89 & 43.85 & 23.10 & 11.93 & 36.93 \\
\textbf{NileChat-MOR} & \textbf{70.67} & 72.56 & 39.94 & \textbf{37.45} & \textbf{30.82} & 29.98 \\
\textbf{NileChat}       & 70.33 & \textbf{72.67} & \textbf{44.37} & 33.89 & 28.67 & \textbf{37.52} \\
\bottomrule
\end{tabular}%
}
\caption{Comparison of the performance of the Egyptian SFT model (\textit{NileChat-EGY}), the Moroccan SFT model (\textit{NileChat-MOR}), and their merged version, \textit{NileChat}, on Belebele (accuracy), Flores (ChrF++), and In-house parallel data (ChrF++).}
\label{tab:merging_diff}
\end{table}


\section{Evaluation Setup}

\subsection{Evaluation Tasks} \label{app:tasks}
\textbf{Full list of the 13 categories of WVS questions.} Economic Values (EcoV); Ethical Values (EthV); Happiness and Wellbeing (HW); Index of Postmaterialism (IP); Perceptions about Science and Technology (PST); Perceptions of Corruption (PC); Perceptions of Migration (PM); Perceptions of Security (PS); Political Culture and Political Regimes (PCPR); Political Interest and Political Participation (PIPP); Religious Values (RV); Social Capital, Trust, and Organizational Membership (SCTOM); and Social Values, Norms, and Stereotypes (SVNS).

\noindent \textbf{The Quality and Validation of Generated Datasets.} We rigorously validated the two evaluation Egyptian Arabic datasets we created using machine translation, namely \textit{EgyMMLU} and \textit{EgyHellaSwag}. A random sample of items from each dataset was rated on two 1–5 scales: (i) \emph{Correctness} (semantic accuracy/faithfulness) and (ii) \emph{Dialectness} (authenticity and naturalness in Egyptian Arabic). Expert human annotator rated 100 randomly sampled items per dataset. To increase statistical power, a state-of-the-art LLM judge (Gemini 2.5 Pro) rated the same 100 items plus an additional 300 (400 total) per dataset. Agreement between human and LLM scores on the 100 shared items yielded an Intraclass Correlation Coefficient (ICC) of \textbf{0.60}, indicating good reliability. Summary scores appear in Table~\ref{tab:validation_quality_check}. These results, together with the inter-evaluator agreement, indicate that the newly created Egyptian benchmarks are of high quality and suitable for reliable model evaluation.

\begin{table}[htbp]
\centering
\begin{tabular}{l l c c c}
\toprule
\textbf{Dataset} & \textbf{Evaluator} & \textbf{Sample Size} & \textbf{Correctness Score} & \textbf{Dialectness Score} \\
\midrule
\multirow{3}{*}{\textbf{EgyMMLU}} & Human & 100 & 3.78 & 3.92 \\
& LLM   & 100 & 3.90 & 4.22 \\
& LLM   & 400 & 4.28 & 4.22 \\
\midrule
\multirow{3}{*}{\textbf{EgyHellaSwag}} & Human & 100 & 3.90 & 4.08 \\
& LLM   & 100 & 4.22 & 4.96 \\
& LLM   & 400 & 4.11 & 4.64 \\
\bottomrule
\end{tabular}
\caption{Translation quality scores (out of 5) for EgyMMLU and EgyHellaSwag along two dimensions: correctness and dialectness. Human raters and an LLM judge rated the same 100 items; the LLM judge additionally rated 300 more items (400 total) to yield more reliable estimates.}
\label{tab:validation_quality_check}
\end{table}

\subsection{Baselines}\label{app:baselines}

We evaluate our model \textit{NileChat} against a set of 17 LLMs that are Arabic-aware; some of these 17 models are also aligned to Arabic dialects. These models are from the following model families: ALLaM \cite{bari2025allam}, Jais \cite{sengupta2023jais}, Atlas-Chat \cite{shang-etal-2025-atlas}, ar-stablelm-2-chat \cite{alyafeai2024arabicstablelmadapting}, Gemma-3 \cite{gemmateam2025gemma3technicalreport}, Qwen-2.5 \cite{qwen2025qwen25technicalreport}, Qwen3 \cite{yang2025qwen3technicalreport} (non-thinking mode), and Llama-3.2 \cite{grattafiori2024llama3herdmodels}. The full list of models, including their corresponding size and release date, are presented in Table \ref{tab:models_params}.

\begin{table}[]
\centering
\begin{tabular}{lll}
\toprule
\multicolumn{1}{c}{\textbf{Model Name}} & \multicolumn{1}{c}{\textbf{Size}} & \multicolumn{1}{c}{\textbf{Release Date}} \\ \hline
\multicolumn{3}{c}{\textit{Less than 7B}} \\
\textbf{Qwen3-1.7B}                     & 1.7                                                & Apr. 2025                                 \\
\textbf{ar-stablelm-2-chat}             & 1.6                                                & Jul. 2024                                 \\
\textbf{Atlas-Chat-2B}                  & 2.6                                                & Sep. 2024                                 \\
\textbf{Llama-3.2-3B-Instruct}          & 3.2                                                & Sep. 2024                                 \\
\textbf{gemma-3-4b-it}                  & 4.3                                                & Mar. 2025                                 \\
\textbf{Qwen3-4B}                       & 4                                                  & Apr. 2025                                 \\
\textbf{NLLB-200-3.3B}                  & 3.3                                                & Jul. 2022                                 \\
\textbf{Qwen2.5-3B-Instruct}            & 3.1                                                  & Sep. 2024                                 \\ \midrule
\multicolumn{3}{c}{\textit{More than 7B}} \\
\textbf{AceGPT-7B-chat}                 & 7                                                  & Dec. 2023                                 \\
\textbf{ALLaM-7B-Instruct}              & 7                                                  & Feb. 2025                                 \\
\textbf{Qwen2.5-7B-Instruct}            & 7.6                                                & Apr. 2025                                 \\
\textbf{Qwen3-8B}                       & 8.2                                                & Apr. 2025                                 \\
\textbf{Atlas-Chat-9B}                  & 9.2                                                & Sep. 2024                                 \\
\textbf{gemma-3-12b-it}                 & 12.2                                               & Mar. 2025                                 \\
\textbf{AceGPT-13B-chat}                & 13                                                 & Dec. 2023                                 \\
\textbf{jais-13b-chat}                  & 13                                                 & Aug. 2023                                 \\ \bottomrule
\end{tabular}
\caption{The LLMs used for comparison against NileChat in this evaluation were selected from a list of Arabic-aware models. Each LLM is listed with its corresponding size (in billion parameters) and release date. We utilized the instruct version for all LLMs except for NLLB, which is a machine translation-specific model.}
\label{tab:models_params}
\end{table}

\section{Full results}\label{app:results}

\begin{table}[]
\centering
\begin{tabular}{ll|cc|cc|cc}
\toprule
\multicolumn{2}{c}{\multirow{2}{*}{\textbf{Model}}} & \multicolumn{2}{c|}{\textbf{MMLU}} & \multicolumn{2}{c|}{\textbf{HellaSwag}} & \multicolumn{2}{c}{\textbf{Belebele}} \\
\cmidrule(lr){3-4} \cmidrule(lr){5-6} \cmidrule(lr){7-8}
\multicolumn{2}{c}{} & \multicolumn{1}{l}{\textbf{EGY}} & \multicolumn{1}{l|}{\textbf{MOR}} & \multicolumn{1}{l}{\textbf{EGY}} & \multicolumn{1}{l|}{\textbf{MOR}} & \multicolumn{1}{l}{\textbf{EGY}} & \multicolumn{1}{l}{\textbf{MOR}} \\
\midrule
\multirow{8}{*}{\rotatebox{90}{\textbf{Less than 7B}}} & \textbf{Qwen3-1.7B}           & 28.53          & 28.53          & 28.07          & 27.33          & 22.89          & 22.89          \\
& \textbf{ar-stablelm-2-chat}     & 39.54          & 38.32          & 34.33          & 33.40          & 24.22          & 22.78          \\
& \textbf{Atlas-Chat-2B}          & 42.65          & 45.06          & 29.62          & 34.78          & 54.67          & 59.00          \\
& \textbf{Llama-3.2-3B-Instruct}  & 31.10          & 30.92          & 28.86          & 28.39          & 49.67          & 40.89          \\
& \textbf{gemma-3-4b-it}          & 46.32          & 46.60          & 34.26          & 32.53          & 61.44          & 52.11          \\
& \textbf{Qwen3-4B}               & 28.59          & 28.52          & 30.21          & 29.53          & 22.89          & 22.89          \\
& \textbf{Qwen2.5-3B-Instruct}    & 35.71          & 37.67          & 31.17          & 29.62          & 61.11          & 44.89          \\
& \textbf{\colorbox{mylightyellow}{NileChat (3B)}}      & \textbf{58.20} & \textbf{58.62} & \textbf{38.29} & \textbf{40.35} & \textbf{78.11} & \textbf{73.78} \\
\midrule
\multirow{8}{*}{\rotatebox{90}{\textbf{More than 7B}}} & \textbf{AceGPT-7B-chat}       & 40.76          & 37.98          & 33.04          & 31.04          & 38.00          & 33.00          \\
& \textbf{ALLaM-7B-Instruct}      & \underline{60.18} & 59.61          & \underline{40.20} & 38.14          & 76.11          & 66.00          \\
& \textbf{Qwen2.5-7B-Instruct}    & 57.70          & 53.51          & 33.79          & 32.28          & 76.67          & 59.44          \\
& \textbf{Qwen3-8B}               & 28.53          & 28.53          & 31.72          & 30.95          & 22.89          & 22.89          \\
& \textbf{Atlas-Chat-9B}          & 57.17          & \underline{60.27} & 34.75          & \underline{44.47} & 78.44          & \underline{79.33} \\
& \textbf{gemma-3-12b-it}         & 59.29          & 56.16          & 40.16          & 37.60          & \underline{80.78} & 73.11          \\
& \textbf{AceGPT-13B-chat}      & 46.48           & 43.65           & 35.15           &  33.21          & 46.33           & 41.11           \\
& \textbf{jais-13b-chat}          & 49.33          & 48.28          & 38.99          & 37.45          & 59.89          & 53.78          \\
\bottomrule
\end{tabular}%
\caption{3-shot performance (accuracy) of models on understanding (MMLU, HellaSwag, and Belebele). Bold values indicate the highest score among models comparable in size to ours (<7B parameters). Underlined values represent the highest score in the entire column, including larger models. Results for zero-shot are presented in Table \ref{tab:unders_cult_results}, Section \ref{sec:results}.}
\label{tab:three_shot_unders}
\end{table}

\begin{sidewaystable}[]
\centering
\resizebox{.98\textwidth}{!}{%
\begin{tabular}{@{}ll*{36}{c}@{}} 
\toprule
\multicolumn{2}{@{}c}{} & \multicolumn{36}{c}{\textbf{Flores Dataset Performance}} \\
\cmidrule(lr){3-38}
& \multicolumn{1}{c}{\textbf{Model}} & \multicolumn{4}{c}{\textbf{MOR $\rightarrow$ EGY}} & \multicolumn{4}{c}{\textbf{MOR $\rightarrow$ ENG}} & \multicolumn{4}{c}{\textbf{MOR $\rightarrow$ FRA}} & \multicolumn{4}{c}{\textbf{EGY $\rightarrow$ MOR}} & \multicolumn{4}{c}{\textbf{EGY $\rightarrow$ ENG}} & \multicolumn{4}{c}{\textbf{ENG $\rightarrow$ EGY}} & \multicolumn{4}{c}{\textbf{ENG $\rightarrow$ EGY}} & \multicolumn{4}{c}{\textbf{FRA $\rightarrow$ EGY}} & \multicolumn{4}{c}{\textbf{Average}} \\
\cmidrule(lr){3-6} \cmidrule(lr){7-10} \cmidrule(lr){11-14} \cmidrule(lr){15-18} \cmidrule(lr){19-22} \cmidrule(lr){23-26} \cmidrule(lr){27-30} \cmidrule(lr){31-34} \cmidrule(lr){35-38}
& & \tiny{sB0} & \tiny{sB4} & \tiny{cF0} & \tiny{cF4} & \tiny{sB0} & \tiny{sB4} & \tiny{cF0} & \tiny{cF4} & \tiny{sB0} & \tiny{sB4} & \tiny{cF0} & \tiny{cF4} & \tiny{sB0} & \tiny{sB4} & \tiny{cF0} & \tiny{cF4} & \tiny{sB0} & \tiny{sB4} & \tiny{cF0} & \tiny{cF4} & \tiny{sB0} & \tiny{sB4} & \tiny{cF0} & \tiny{cF4} & \tiny{sB0} & \tiny{sB4} & \tiny{cF0} & \tiny{cF4} & \tiny{sB0} & \tiny{sB4} & \tiny{cF0} & \tiny{cF4} & \tiny{sB0} & \tiny{sB4} & \tiny{cF0} & \tiny{cF4} \\
\midrule
\multirow{9}{*}{\rotatebox[origin=c]{90}{\textbf{Less than 7B}}} & \textbf{Qwen3-1.7B} & 19.49 & 19.80 & 36.32 & 36.45 & 14.76 & 15.45 & 40.21 & 40.79 & 12.17 & 12.90 & 35.08 & 35.83 & 19.38 & 19.31 & 36.23 & 36.22 & 19.65 & 20.48 & 46.49 & 47.05 & 7.32 & 7.77 & 22.02 & 23.42 & 10.02 & 10.62 & 25.98 & 27.55 & 5.98 & 6.63 & 19.60 & 21.69 & 13.59 & 14.12 & 32.74 & 33.62 \\
& \textbf{ar-stablelm-2-chat} & 18.29 & 18.57 & 34.31 & 34.46 & 7.36 & 7.03 & 28.97 & 28.47 & 3.51 & 2.85 & 19.36 & 18.08 & 14.16 & 16.35 & 29.25 & 33.32 & 8.05 & 7.93 & 31.12 & 30.67 & 3.25 & 11.51 & 8.22 & 29.25 & 10.41 & 18.68 & 22.89 & 35.77 & 3.81 & 9.55 & 9.06 & 26.13 & 8.60 & 11.56 & 22.90 & 29.52 \\
& \textbf{Atlas-Chat-2B} & 19.67 & 20.03 & 36.67 & 36.90 & 24.64 & 26.18 & 49.36 & 50.32 & 19.03 & 20.18 & 41.53 & 42.49 & 19.35 & 19.88 & 36.88 & \textbf{37.65} & 23.44 & 25.41 & 49.17 & 50.28 & 12.06 & 14.89 & 29.66 & 33.48 & 10.74 & 15.16 & 27.45 & 33.04 & 8.80 & 11.72 & 25.73 & 29.77 & 17.22 & 19.18 & 37.06 & 39.24 \\
& \textbf{Llama-3.2-3B-Instruct} & 18.75 & 10.79 & 35.15 & 21.71 & 15.71 & 0.08 & 38.75 & 0.85 & 12.16 & 0.54 & 29.55 & 3.40 & 17.67 & 14.00 & 34.59 & 28.64 & 20.89 & 0.06 & 45.19 & 0.64 & 5.86 & 0.91 & 16.92 & 5.21 & 9.74 & 0.77 & 24.54 & 5.29 & 3.92 & 1.12 & 15.14 & 5.64 & 13.09 & 3.53 & 29.98 & 8.92 \\
& \textbf{gemma-3-4b-it} & 16.89 & 21.11 & 33.00 & 37.78 & 7.99 & 24.48 & 17.81 & 49.50 & 5.52 & \textbf{23.80} & 11.28 & \textbf{46.04} & 12.86 & 16.31 & 29.10 & 33.91 & 12.05 & 29.04 & 24.62 & 54.24 & 1.88 & 13.05 & 4.85 & 31.03 & 1.65 & 20.13 & 3.24 & 37.60 & 0.93 & 10.56 & 3.01 & 28.17 & 7.47 & 19.81 & 15.86 & 39.78 \\
& \textbf{Qwen3-4B} & 20.41 & 20.57 & 37.17 & 37.15 & 18.98 & 20.67 & 44.68 & 46.07 & 17.30 & 18.39 & 39.96 & 41.33 & 16.06 & 18.86 & 31.90 & 35.84 & 23.99 & 25.75 & 50.47 & 51.83 & 11.43 & 11.30 & 27.93 & 27.98 & 15.46 & 15.83 & 33.00 & 33.20 & 7.44 & 9.54 & 21.46 & 25.90 & 16.38 & 17.62 & 35.82 & 37.41 \\
& \textbf{NLLB-200-3.3B} & 20.92 & - & 38.74 & - & \textbf{30.89} & - & \textbf{53.64} & - & \textbf{27.89} & - & \textbf{48.69} & - & 17.06 & - & 35.14 & - & \textbf{\underline{34.62}} & - & \underline{\textbf{58.07}} & - & \textbf{\underline{17.46}} & - & \textbf{34.89} & - & \textbf{\underline{26.93}} & - & \underline{\textbf{43.86}} & - & 11.58 & - & 28.90 & - & \textbf{\underline{23.42}} & - & \textbf{42.74} & - \\
& \textbf{Qwen2.5-3B-Instruct} & 18.51 & 19.71 & 35.20 & 36.32 & 18.61 & 19.83 & 44.22 & 44.99 & 15.00 & 16.22 & 38.56 & 39.45 & 16.98 & 17.90 & 34.15 & 34.91 & 24.07 & 25.08 & 50.45 & 50.99 & 9.36 & 8.48 & 25.17 & 25.14 & 11.78 & 11.43 & 28.05 & 28.44 & 7.47 & 7.18 & 22.79 & 23.32 & 15.22 & 15.73 & 34.83 & 35.45 \\
& \textbf{\colorbox{mylightyellow}{NileChat}} & \textbf{23.81} & \textbf{22.68} & \textbf{40.25} & \textbf{39.25} & 28.81 & \textbf{29.50} & 52.98 & \textbf{53.43} & 24.05 & 23.49 & 46.30 & 45.56 & \textbf{\underline{20.90}} & \underline{\textbf{20.51}} & \underline{\textbf{38.22}} & 37.64 & 30.58 & \textbf{31.90} & 54.53 & \textbf{56.19} & 15.15 & \textbf{17.90} & 30.76 & \textbf{36.00} & 23.39 & \textbf{25.03} & 40.37 & \textbf{41.94} & \textbf{13.18} & \textbf{14.91} & \textbf{29.61} & \textbf{32.52} & 22.49 & \textbf{23.24} & 41.63 & \textbf{42.82} \\
\midrule
\multirow{8}{*}{\rotatebox[origin=c]{90}{\textbf{More than 7B}}} & \textbf{AceGPT-7B-chat} & 19.92 & 18.93 & 36.91 & 35.72 & 17.67 & 20.04 & 44.09 & 45.71 & 14.78 & 15.92 & 38.08 & 38.87 & 19.11 & 17.99 & 36.04 & 35.24 & 23.11 & 25.81 & 50.28 & 51.96 & 8.93 & 12.29 & 21.59 & 29.72 & 16.11 & 18.25 & 33.32 & 35.56 & 5.95 & 10.28 & 16.65 & 27.07 & 15.70 & 17.44 & 34.62 & 37.48 \\
& \textbf{ALLaM-7B-Instruct} & \underline{24.63} & \underline{25.12} & \underline{40.93} & \underline{41.39} & 26.02 & 28.13 & 51.65 & 52.92 & 18.91 & 20.80 & 42.64 & 44.04 & 19.85 & 19.53 & 37.32 & 37.00 & 29.63 & 31.82 & 55.53 & 56.75 & 14.63 & 16.56 & 32.38 & 34.79 & 23.18 & \underline{25.93} & 39.85 & \underline{42.70} & 13.17 & 14.09 & 30.75 & 32.15 & 21.25 & 22.75 & 41.38 & 42.72 \\
& \textbf{Qwen2.5-7B-Instruct} & 16.28 & 18.87 & 32.19 & 35.08 & 21.59 & 23.62 & 47.28 & 48.55 & 18.98 & 19.96 & 42.12 & 42.93 & 13.24 & 16.09 & 29.34 & 32.91 & 26.39 & 27.83 & 52.72 & 53.74 & 9.73 & 11.19 & 25.32 & 28.28 & 12.54 & 15.00 & 27.80 & 31.34 & 7.73 & 9.75 & 22.71 & 26.46 & 15.81 & 17.79 & 34.93 & 37.41 \\
& \textbf{Qwen3-8B} & 21.10 & 21.41 & 37.84 & 38.13 & 22.41 & 24.33 & 48.04 & 49.51 & 20.46 & 21.91 & 43.17 & 44.32 & 17.82 & 18.35 & 34.63 & 35.40 & 27.30 & 29.36 & 53.32 & 54.77 & 13.60 & 12.56 & 31.05 & 29.63 & 18.95 & 18.83 & 36.79 & 36.72 & 10.16 & 10.27 & 26.33 & 26.86 & 18.98 & 19.63 & 38.90 & 39.42 \\
& \textbf{Atlas-Chat-9B} & 19.44 & 20.52 & 36.32 & 37.29 & \underline{31.29} & \underline{32.75} & \underline{54.19} & \underline{55.61} & \underline{28.14} & 28.82 & \underline{49.35} & 49.85 & 18.86 & 20.36 & 36.78 & \underline{37.91} & 30.98 & \underline{32.83} & 55.04 & 56.07 & 17.29 & \underline{18.91} & \underline{35.83} & \underline{37.42} & 16.95 & 19.69 & 34.35 & 37.25 & \underline{14.53} & \underline{15.59} & \underline{32.76} & \underline{33.92} & 22.19 & \underline{23.68} & \underline{41.83} & 43.16 \\
& \textbf{gemma-3-12b-it} & 19.67 & 24.48 & 36.28 & 41.24 & 23.68 & 30.29 & 47.75 & 54.47 & 15.28 & \underline{29.99} & 26.64 & \underline{51.36} & 10.73 & 17.30 & 24.98 & 36.28 & 27.38 & 32.71 & 53.46 & \underline{57.42} & 2.51 & 14.60 & 6.56 & 34.00 & 6.34 & 24.27 & 11.17 & 41.69 & 1.45 & 12.47 & 3.70 & 31.30 & 13.38 & 23.26 & 26.32 & \underline{43.47} \\
& \textbf{AceGPT-13B-chat} & 20.14 & 19.58 & 37.04 & 36.57 & 20.82 & 24.30 & 46.12 & 49.11 & 18.58 & 20.52 & 40.45 & 42.77 & 19.54 & 17.99 & 36.62 & 35.50 & 26.08 & 29.14 & 52.52 & 54.71 & 13.02 & 13.42 & 29.23 & 31.52 & 18.81 & 20.18 & 36.04 & 37.93 & 9.51 & 11.72 & 23.19 & 29.39 & 18.31 & 19.61 & 37.65 & 39.69 \\
& \textbf{jais-13b-chat} & 13.50 & 20.28 & 26.36 & 37.46 & 19.73 & 27.40 & 37.81 & 51.80 & 18.14 & 18.40 & 39.16 & 40.95 & 8.90 & 13.42 & 22.22 & 31.25 & 22.63 & 30.07 & 41.73 & 54.56 & 2.46 & 8.55 & 6.08 & 25.73 & 4.09 & 18.77 & 9.56 & 36.94 & 1.50 & 7.26 & 3.25 & 23.70 & 11.37 & 18.02 & 23.27 & 37.80 \\
\midrule[\heavyrulewidth]
\multicolumn{2}{@{}c}{} & \multicolumn{36}{c}{\textbf{In-House Dataset Performance}} \\
\cmidrule(lr){3-38}
& \multicolumn{1}{c}{\textbf{Model}} & \multicolumn{4}{c}{\textbf{EGY $\rightarrow$ ENG}} & \multicolumn{4}{c}{\textbf{EGY $\rightarrow$ MSA}} & \multicolumn{4}{c}{\textbf{ENG $\rightarrow$ EGY}} & \multicolumn{4}{c}{\textbf{ENG $\rightarrow$ MOR}} & \multicolumn{4}{c}{\textbf{MOR $\rightarrow$ EGY}} & \multicolumn{4}{c}{\textbf{MOR $\rightarrow$ MSA}} & \multicolumn{4}{c}{\textbf{MSA $\rightarrow$ EGY}} & \multicolumn{4}{c}{\textbf{MSA $\rightarrow$ MOR}} & \multicolumn{4}{c}{\textbf{Average}} \\
\cmidrule(lr){3-6} \cmidrule(lr){7-10} \cmidrule(lr){11-14} \cmidrule(lr){15-18} \cmidrule(lr){19-22} \cmidrule(lr){23-26} \cmidrule(lr){27-30} \cmidrule(lr){31-34} \cmidrule(lr){35-38}
& & \tiny{sB0} & \tiny{sB4} & \tiny{cF0} & \tiny{cF4} & \tiny{sB0} & \tiny{sB4} & \tiny{cF0} & \tiny{cF4} & \tiny{sB0} & \tiny{sB4} & \tiny{cF0} & \tiny{cF4} & \tiny{sB0} & \tiny{sB4} & \tiny{cF0} & \tiny{cF4} & \tiny{sB0} & \tiny{sB4} & \tiny{cF0} & \tiny{cF4} & \tiny{sB0} & \tiny{sB4} & \tiny{cF0} & \tiny{cF4} & \tiny{sB0} & \tiny{sB4} & \tiny{cF0} & \tiny{cF4} & \tiny{sB0} & \tiny{sB4} & \tiny{cF0} & \tiny{cF4} & \tiny{sB0} & \tiny{sB4} & \tiny{cF0} & \tiny{cF4} \\
\midrule
\multirow{9}{*}{\rotatebox[origin=c]{90}{\textbf{Less than 7B}}} & \textbf{Qwen3-1.7B} & 12.14 & 13.14 & 31.46 & 31.98 & 19.11 & 20.32 & 34.24 & 35.09 & 4.63 & 4.96 & 15.10 & 15.43 & 2.52 & 2.64 & 11.14 & 10.99 & 6.19 & 6.12 & 22.58 & 23.01 & 6.45 & 6.04 & 19.25 & 18.12 & 18.20 & 18.26 & 34.20 & 33.86 & 6.20 & 6.07 & 18.38 & 17.94 & 9.43 & 9.69 & 23.29 & 23.30 \\
& \textbf{ar-stablelm-2-chat} & 7.89 & 6.74 & 21.92 & 20.28 & 14.56 & 13.71 & 27.71 & 24.23 & 3.57 & 10.99 & 7.50 & 22.25 & 0.65 & 3.69 & 1.45 & 14.61 & 6.95 & 6.15 & 20.63 & 19.05 & 8.50 & 7.28 & 21.32 & 18.59 & 14.89 & 14.28 & 28.14 & 27.20 & 5.19 & 3.49 & 15.82 & 13.12 & 7.77 & 8.29 & 18.06 & 19.92 \\
& \textbf{Atlas-Chat-2B} & 13.52 & 16.21 & 32.04 & 34.25 & 15.51 & 18.19 & 30.39 & 32.34 & 3.67 & 5.15 & 15.29 & 17.15 & 6.81 & 8.40 & 21.76 & 23.79 & 14.96 & 15.85 & 34.83 & 35.82 & 12.12 & 12.76 & 28.38 & 28.48 & 7.05 & 11.91 & 20.72 & 26.75 & 8.86 & 10.11 & 24.35 & 25.20 & 10.31 & 12.32 & 25.97 & 27.97 \\
& \textbf{Llama-3.2-3B-Instruct} & 12.97 & 0.38 & 28.40 & 1.37 & 14.24 & 4.80 & 25.90 & 8.81 & 5.60 & 2.41 & 16.44 & 8.16 & 2.03 & 0.98 & 10.14 & 5.63 & 4.74 & 0.23 & 17.03 & 1.01 & 5.00 & 1.98 & 14.55 & 6.79 & 15.73 & 7.97 & 31.80 & 18.17 & 4.30 & 2.46 & 14.68 & 11.25 & 8.08 & 2.65 & 19.87 & 7.65 \\
& \textbf{gemma-3-4b-it} & 16.24 & 22.91 & 32.96 & 43.72 & 17.53 & 26.98 & 30.61 & 42.52 & 0.52 & 13.66 & 1.61 & 28.41 & 0.49 & 5.67 & 1.90 & 18.59 & 6.10 & 12.65 & 16.81 & 33.20 & 4.41 & 12.67 & 11.22 & 28.30 & 5.50 & 19.30 & 11.35 & 35.56 & 0.72 & 6.16 & 2.13 & 18.92 & 6.44 & 15.00 & 13.57 & 31.15 \\
& \textbf{Qwen3-4B} & 19.21 & 18.42 & 39.11 & 38.95 & 22.22 & 23.63 & 37.84 & 38.93 & 8.19 & 8.16 & 20.04 & 20.48 & 3.41 & 3.40 & 13.35 & 13.57 & 9.18 & 8.54 & 26.74 & 26.42 & 7.86 & 8.51 & 21.12 & 21.64 & 17.98 & 19.72 & 33.83 & 35.71 & 5.47 & 6.15 & 16.43 & 18.51 & 11.69 & 12.07 & 26.06 & 26.78 \\
& \textbf{NLLB-200-3.3B} & 21.48 & - & 41.14 & - & 16.33 & - & 34.36 & - & 14.20 & - & 29.78 & - & 6.28 & - & 19.86 & - & 14.12 & - & 32.44 & - & 8.75 & - & 26.11 & - & 19.34 & - & 36.68 & - & 8.70 & - & 23.10 & - & 13.65 & - & 30.43 & - \\
& \textbf{Qwen2.5-3B-Instruct} & 19.44 & 19.38 & 39.84 & 39.59 & 19.04 & 20.48 & 34.07 & 34.75 & 6.34 & 6.58 & 17.94 & 18.97 & 3.38 & 3.65 & 13.01 & 14.36 & 8.15 & 9.49 & 26.02 & 27.75 & 7.52 & 7.68 & 20.87 & 20.82 & 13.47 & 16.82 & 28.91 & 33.21 & 5.00 & 5.40 & 17.04 & 17.88 & 10.29 & 11.19 & 24.71 & 25.92 \\
& \textbf{\colorbox{mylightyellow}{NileChat}} & \textbf{26.24} & \textbf{27.49} & \textbf{46.46} & \textbf{48.10} & \textbf{26.77} & \textbf{31.65} & \textbf{42.52} & \textbf{46.40} & \textbf{\underline{17.67}} & \underline{\textbf{18.92}} & \underline{\textbf{33.37}} & \underline{\textbf{34.87}} & \textbf{\underline{10.74}} & \underline{\textbf{11.72}} & \underline{\textbf{25.07}} & \underline{\textbf{26.84}} & \textbf{\underline{19.22}} & \underline{\textbf{20.47}} & \underline{\textbf{39.83}} & \underline{\textbf{41.46}} & \textbf{17.55} & \textbf{19.28} & \textbf{34.23} & \underline{\textbf{36.55}} & \textbf{\underline{26.37}} & \underline{\textbf{29.54}} & \underline{\textbf{42.90}} & \underline{\textbf{46.68}} & \textbf{\underline{13.94}} & \underline{\textbf{15.41}} & \underline{\textbf{29.37}} & \underline{\textbf{30.75}} & \textbf{\underline{19.81}} & \underline{\textbf{21.81}} & \underline{\textbf{36.72}} & \underline{\textbf{38.95}} \\
\midrule
\multirow{8}{*}{\rotatebox[origin=c]{90}{\textbf{More than 7B}}} & \textbf{AceGPT-7B-chat} & 19.27 & 21.44 & 39.36 & 40.93 & 20.94 & 21.99 & 35.99 & 37.51 & 9.24 & 10.57 & 22.39 & 25.25 & 3.36 & 3.69 & 12.76 & 15.12 & 7.80 & 9.95 & 26.08 & 27.97 & 7.15 & 8.23 & 20.20 & 22.17 & 20.22 & 20.47 & 37.14 & 37.05 & 6.53 & 6.00 & 19.31 & 18.98 & 11.81 & 12.79 & 26.65 & 28.12 \\
& \textbf{ALLaM-7B-Instruct} & \underline{26.31} & \underline{28.98} & \underline{48.32} & \underline{49.57} & \underline{32.49} & \underline{32.85} & \underline{48.22} & 47.93 & 16.69 & 17.78 & 32.85 & 34.12 & 8.05 & 9.32 & 22.65 & 24.47 & 18.61 & 19.97 & 39.44 & 40.36 & \underline{18.41} & \underline{19.60} & \underline{34.86} & 36.02 & 23.28 & 26.15 & 41.09 & 43.79 & 10.28 & 11.13 & 25.95 & 26.49 & 19.26 & 20.72 & 36.67 & 37.84 \\
& \textbf{Qwen2.5-7B-Instruct} & 22.70 & 21.40 & 43.05 & 42.34 & 19.14 & 22.90 & 33.08 & 36.71 & 7.65 & 7.91 & 19.04 & 20.16 & 3.65 & 4.50 & 13.87 & 16.04 & 9.87 & 10.55 & 28.38 & 29.17 & 7.72 & 9.05 & 20.84 & 22.79 & 13.21 & 16.00 & 28.10 & 31.91 & 4.55 & 5.74 & 15.67 & 18.44 & 11.06 & 12.26 & 25.25 & 27.20 \\
& \textbf{Qwen3-8B} & 22.21 & 23.62 & 42.36 & 44.34 & 26.06 & 27.35 & 41.28 & 42.38 & 9.04 & 9.52 & 22.25 & 23.18 & 3.67 & 4.50 & 14.64 & 15.68 & 10.23 & 11.03 & 28.90 & 29.63 & 8.30 & 9.16 & 21.79 & 23.18 & 17.71 & 19.49 & 33.99 & 35.90 & 5.79 & 6.12 & 17.83 & 18.83 & 12.88 & 13.85 & 27.88 & 29.14 \\
& \textbf{Atlas-Chat-9B} & 18.98 & 21.19 & 38.91 & 41.38 & 15.72 & 20.68 & 31.29 & 35.84 & 4.53 & 7.74 & 17.12 & 21.46 & 7.19 & 10.60 & 22.68 & 26.51 & 16.77 & 19.19 & 37.84 & 40.15 & 13.70 & 15.38 & 29.86 & 31.29 & 6.19 & 12.05 & 19.20 & 27.01 & 8.16 & 12.53 & 23.70 & 28.12 & 11.41 & 14.92 & 27.58 & 31.47 \\
& \textbf{gemma-3-12b-it} & 23.12 & 28.51 & 46.31 & 48.58 & 25.91 & 32.60 & 40.06 & \underline{48.67} & 6.59 & 17.70 & 14.15 & 34.10 & 1.83 & 7.95 & 6.29 & 23.15 & 12.49 & 18.35 & 33.40 & 39.90 & 12.26 & 17.54 & 23.93 & 34.51 & 9.14 & 26.48 & 18.35 & 43.51 & 3.06 & 10.73 & 8.83 & 26.07 & 11.80 & 19.98 & 23.91 & 37.31 \\
& \textbf{AceGPT-13B-chat} & 22.36 & 25.37 & 42.53 & 45.47 & 24.67 & 27.42 & 40.60 & 42.11 & 10.76 & 12.56 & 25.54 & 27.49 & 4.27 & 5.17 & 15.26 & 17.90 & 10.56 & 12.24 & 28.12 & 31.28 & 8.47 & 10.30 & 22.28 & 24.78 & 20.32 & 22.83 & 37.61 & 39.75 & 6.86 & 7.31 & 20.34 & 20.72 & 13.53 & 15.40 & 29.04 & 31.19 \\
& \textbf{jais-13b-chat} & 16.71 & 24.94 & 33.16 & 43.30 & 21.68 & 28.15 & 36.23 & 43.52 & 6.45 & 14.19 & 14.78 & 30.35 & 2.47 & 6.35 & 7.42 & 20.10 & 12.06 & 18.56 & 26.34 & 38.44 & 12.89 & 18.74 & 28.36 & 34.79 & 15.20 & 22.74 & 30.81 & 39.83 & 5.57 & 8.83 & 16.16 & 23.75 & 11.63 & 17.81 & 24.16 & 34.26 \\
\bottomrule
\end{tabular}%
}
\caption{Detailed translation evaluation results on Flores and In-House datasets. Metrics: spBLEU (sB) and ChrF++ (cF) for 0-shot (0) and 4-shot (4) scenarios. Bold values highlight the top score among models with fewer than 7 billion parameters. Underlined values indicate the highest score overall in each column.}
\label{tab:combined_flores_inhouse_stacked}
\end{sidewaystable}

\section{Prompts} \label{app:prompts}

The provided figures showcase diverse prompts for language models targeting low-resource languages. Figure \ref{fig:query1_listing} translates English educational content into conversational dialectal Arabic, while Figure \ref{fig:query2_listing} converts dialectal Arabic script to Arabizi. English instructions are translated to dialectal Arabic using the prompt in Figure \ref{fig:query3_listing}. For content generation, Figure \ref{fig:query4_listing} guides the model to create culturally relevant dialectal Arabic text based on a given persona and context. Figure \ref{fig:query5_listing} focuses on summarizing detailed persona descriptions concisely. Finally, Figure \ref{fig:query6_listing} instructs an LLM to generate practical dialectal Arabic question-answer pairs in JSON format from provided text.

\begin{figure}[H]
    \centering
\begin{lstlisting}[breaklines=true]
Translate the following text from English to Egyptian Arabic. Ensure that all words are in Egyptian Arabic, and do not use any Modern Standard Arabic (MSA). Keep the translation casual, conversational, and reflective of how Egyptians would naturally speak in everyday situations. Avoid any formal or classical language structures. Translate only the input paragraph and don't add anything else in your output.
English: {English_text}
\end{lstlisting}
    \caption{The translation prompt used with a teacher model to convert English educational pre-training data to a low-resource target language. The placeholder '\{English\_text\}' represents the input English text.}
    \label{fig:query1_listing}
\end{figure}

\begin{figure}[H]
    \centering
    \begin{lstlisting}[breaklines=true]
Write the following Moroccan dialectal Arabic text in Moroccan Arabizi. Ensure that all words are written in Moroccan Arabizi. Keep the text casual, conversational, and reflective of how Moroccans would naturally write in everyday situations using Arabizi. Translate only the content keys in the following JSON, and output a json of the same format:
{JSON_OBJECT}
    \end{lstlisting}
    \caption{The prompt used with our teacher LLM to convert dialectal Arabic text written in Arabic script into Arabizi. The placeholder \{JSON\_OBJECT\} represents the input text formatted as a JSON object.}
    \label{fig:query2_listing} 
\end{figure}

\begin{figure}[H]
    \centering
\begin{lstlisting}[breaklines=true]
Translate the following text from English to Moroccan Arabic. Ensure that all words are in Moroccan Arabic, and do not use any Modern Standard Arabic (MSA). Keep the translation casual, conversational, and reflective of how Moroccans would naturally speak in everyday situations. Avoid any formal or classical language structures. Translate only the content keys in the following JSON, and output a json of the same format:
{JSON_OBJECT}
\end{lstlisting}
    \caption{The translation prompt used with a teacher model to convert SmolTalk and TULU  instructions data to a low-resource target language. The placeholder '\{JSON\_OBJECT\}' represents the input text.}
    \label{fig:query3_listing}
\end{figure}

\begin{figure}[H]
    \centering
\begin{lstlisting}[breaklines=true]
Act as the following person: {persona_description} Act like you are {person_Name} and write a {text_genre} in Egyptian dialect, using colloquial Arabic script as spoken in Egypt and not Modern Standard Arabic (MSA). Use this context and use the information provided in it while writing the {text_genre}:
{context}
Make sure to follow these conditions: 
1. Rely on the provided context when writing the {text_genre}.
2. Ensure that the written {text_genre} reflects the cultural background, values, and worldview of {person_Name}.
3. Don't write the persona's description. I want you to focus only on the provided context when writing while reflecting the perosna's background.
Note: Ensure that all words are in Egyptian Arabic, and do not use any Modern Standard Arabic (MSA). Keep the translation casual, conversational, and reflective of how Egyptians would naturally speak in everyday situations.
\end{lstlisting}
    \caption{Prompt for generating culturally and values-aware text genres in low-resource languages\texttt{,} given a local persona description and a local cultural concept. The placeholders \texttt{\{persona\_description\}}\texttt{,} \texttt{\{text\_genre\}}\texttt{,} and \texttt{\{context\}} represent the persona description\texttt{,} the intended text genre to generate\texttt{,} and the cultural concept text\texttt{,} respectively.}
    \label{fig:query4_listing}
\end{figure}

\begin{figure}[H]
    \centering
\begin{lstlisting}[breaklines=true]
I have the following persona description, I want you to write it in a concise manner keeping all the information, the output should be plain text, make sure to include all values, morals, and culture of the persona:
{PERSONA_DESCRIPTION}
\end{lstlisting}
    \caption{Prompt for generating concise persona descriptions with a LLM utilizing a comprehensive description of a specific persona extracted from the WVS. The placeholder '\{PERSONA\_DESCRIPTION\}' represents the input persona description.}
    \label{fig:query5_listing}
\end{figure}

\begin{figure}[H]
    \centering
    
\begin{lstlisting}[breaklines=true, basicstyle=\scriptsize]
You are an expert in Moroccan culture and language, with native-level fluency in Moroccan Arabic dialect written in Arabic script. Your task is to transform detailed articles or information (like recipes, historical accounts, cultural traditions, etc.) into practical, useful instruction-response pairs in authentic Moroccan Arabic dialect expressed in Arabic script.

## Task Description:
1. I will provide you with raw text content (like cooking recipes, cultural traditions, historical information, etc.)
2. Transform this content into practical instruction-response pairs where:
   - Instructions ask how to do something, how something works, or how to understand something
   - Responses provide thorough, practical explanations in Moroccan dialect
3. Focus on making these interactions useful for someone wanting to learn practical information
4. Output the results in structured JSON format

## Instruction Format Guidelines:
- For procedural content (recipes, crafts, etc.): Create "how to" questions
  - Example: {example_1}
- For historical/factual content: Create "why" questions
  - Example: {example_2}
- For cultural practices: Create "what is" or "how do we celebrate" questions
  - Example: {example_3}

## Response Format Guidelines:
- Make responses detailed and practical
- Include specific steps for procedural content
- Use authentic Moroccan vocabulary and expressions
- Structure responses in a logical order 
- For recipes or procedures, list steps clearly in the order they should be performed
- Include tips and warnings where appropriate
- Keep the tone conversational and helpful, as if explaining to a friend

## Language Guidelines:
- Use authentic Moroccan Arabic dialect written in Arabic script (not transliteration)
- Include typical Moroccan cooking/cultural terminology and expressions
- Use language as it would naturally be spoken, not literary Arabic
- Include common Moroccan filler words and expressions where natural
- For cooking terms or specialized vocabulary, use the actual terms Moroccans use.

## Output Format:
The output should be valid JSON with the following structure:
```json
{
  "instruction_response_pairs": [
    {
      "instruction": "[practical instruction in Moroccan dialect]",
      "response": "[detailed practical response in Moroccan dialect]"
    },
    {
      "instruction": "[another practical instruction]",
      "response": "[another detailed response]"
    }
  ]
}
```
## Instructions:
1. Generate at least 3-5 practical instruction-response pairs based on the content I provide.
2. Make each instruction focus on a practical aspect (how-to, ingredients, common mistakes, variations, etc.).
3. Ensure responses are detailed enough to be practically useful.
4. The resulting JSON must be properly formatted and valid.
5. For specialized terms or techniques, use the authentic Moroccan terms.
6. When dealing with recipes or procedures, make sure the steps are in logical order.
7. If the article contains cultural context or historical information, include some of this in the responses to provide depth.
8. If the article does not have any relevent information to the Moroccan culture then return an empty list.

Now, I'll provide you with content, and you'll transform it into practical instruction-response pairs in Moroccan dialect as specified above.
{Text}
\end{lstlisting}

    \caption{Prompt for generating synthetic cultural dialectal question-answer pairs using a teacher model given raw text describing a local cultural heritage concept. The placeholder \{Text\} represents the raw text that is used to generate question-answer pairs from it.}
    \label{fig:query6_listing}
\end{figure}

\section{Samples of texts generated from NileChat}

\begin{table*}[t]
\centering 
\includegraphics[width=0.95\linewidth, trim={0pt 0pt 0pt 0pt}, clip]{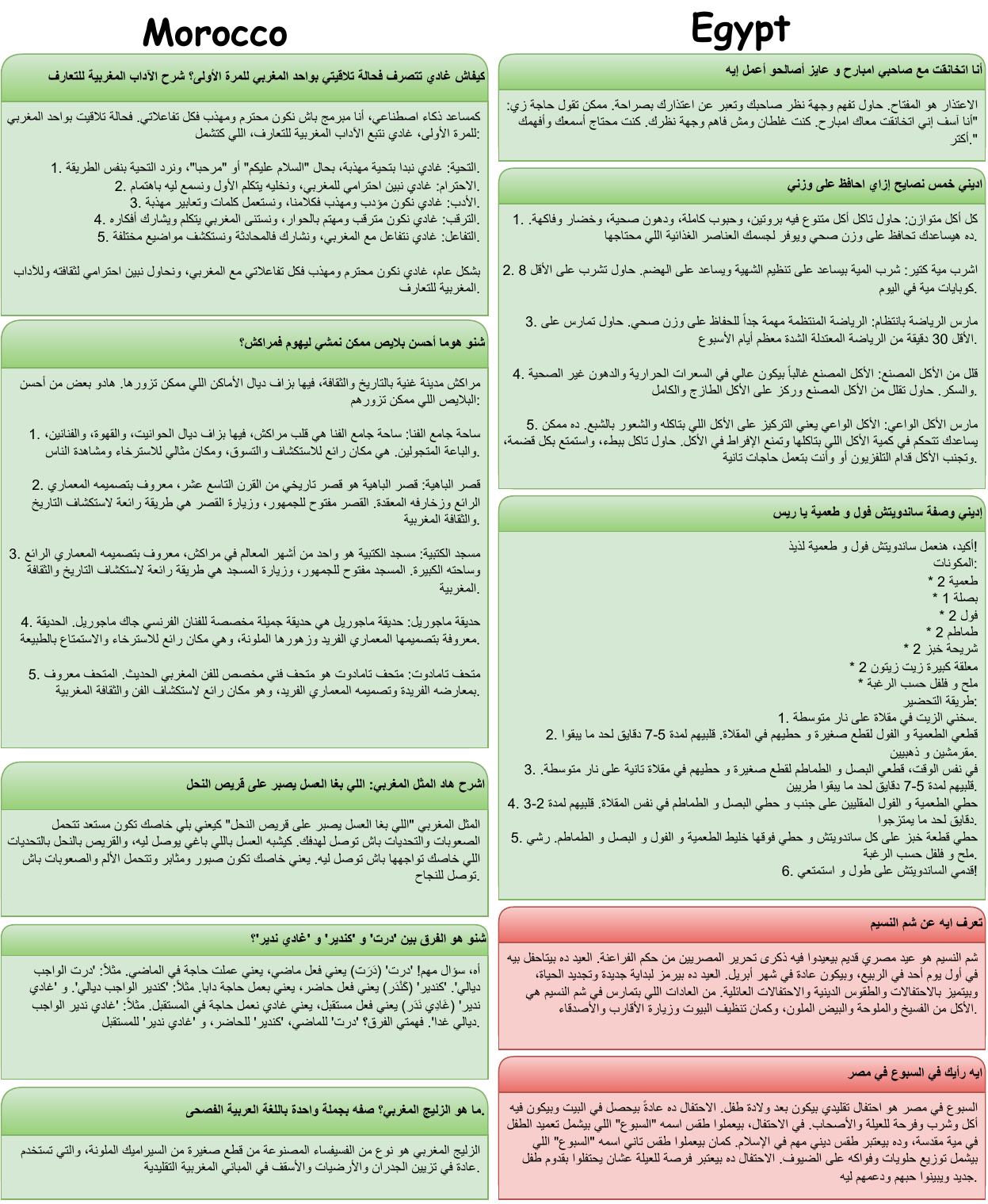}
\caption{Sample responses from \textit{NileChat} to prompts in Egyptian and Moroccan dialects, covering general and local cultural knowledge. Samples with green background color represent samples with correct responses, samples with red background color represent samples with not accurate answers.}
\label{tab:nilechat_samples}
\end{table*}

\end{document}